\documentclass[conference]{IEEEtran}

\usepackage[T1]{fontenc}
\usepackage[utf8]{inputenc}
\usepackage{cite}
\usepackage{amsmath,amssymb,amsfonts}
\usepackage{graphicx}
\usepackage{booktabs}
\usepackage{array}
\usepackage{multirow}
\usepackage{url}
\usepackage{xcolor}
\usepackage{placeins}
\usepackage{hyperref}
\usepackage{enumitem}
\usepackage{algorithm}
\usepackage{algpseudocode}
\usepackage{orcidlink}
\usepackage{todonotes}
\usepackage{tabularx}
\usepackage{svg}
\usepackage{longtable}
\usepackage{ragged2e}
\usepackage{subcaption}
\usepackage{listings}
\usepackage{tikz}
\newcolumntype{L}[1]{>{\RaggedRight\arraybackslash}p{#1}}
\hypersetup{
  colorlinks=true,
  linkcolor=black,
  citecolor=black,
  urlcolor=blue
}
\usetikzlibrary{shapes.geometric, arrows.meta, positioning, fit, backgrounds, calc}

\newcommand{\mesocort}{Mesocort}

\title{Toward Fast Human Reaction Modeling\\from High-Frequency Digital Interaction Data\\}

\author{%
\IEEEauthorblockN{%
Emanuel Dopater\IEEEauthorrefmark{1},
Daniel Gavlas\IEEEauthorrefmark{1},
Luk\'a\v{s} Jochymek\IEEEauthorrefmark{1},
Kannan Paul\IEEEauthorrefmark{1},
Miloslav Szczypka\IEEEauthorrefmark{1},
David Voj\'a\v{c}ek\IEEEauthorrefmark{1},
Josef \v{Z}\'a\v{c}ek\IEEEauthorrefmark{1},\\
Jan Plucar\,\orcidlink{0000-0001-7620-9941}\IEEEauthorrefmark{2},
Radek Svoboda\,\orcidlink{0000-0002-8017-8911}\IEEEauthorrefmark{2},
Ivan Zelinka\,\orcidlink{0000-0002-3858-7340}\IEEEauthorrefmark{2}}

\IEEEauthorblockA{\IEEEauthorrefmark{1}%
Mesocort AI\\
Tov\'arn\'i 589, 709 00 Ostrava 9, Czechia\\
\{emanuel.dopater, lukas.jochymek, kannan, miloslav.szczypka, david.vojacek, josef.zacek\}@mesocort.ai}

\IEEEauthorblockA{\IEEEauthorrefmark{2}%
Faculty of Electrical Engineering and Computer Science\\
V\v{S}B--Technical University of Ostrava\\
17.\ listopadu 2172/15, 70833 Ostrava, Czechia\\
\{jan.plucar, radek.svoboda, ivan.zelinka\}@vsb.cz}
}

\begin{document}
\maketitle

\begin{abstract}
We ask whether limbic-like behavioral signatures, fast salience-driven responses preceding deliberation, can be detected and modeled from high-resolution Counter-Strike~2 telemetry. As our first work in this area, the paper is primarily exploratory: it brings several ideas together into a single framework for studying rapid reactions, response latency, and decision-making under pressure, and some parts are necessarily preliminary. The central contribution is an Ego-Knowledge representation, a leak-reduced reconstruction of what each player could plausibly perceive rather than omniscient replay truth, together with a millisecond-precision visibility engine, a reaction-centered extraction pipeline, and an intracranially motivated 60--120~ms operational research window. We derive coarse fight-or-flight labels for a mixed dataset of visibility-triggered and damage-triggered events and train two telemetry-native models using information available at or before the trigger: EMBER (gradient boosting) and ORBIT (a custom transformer). Back-damage events provide a constrained setting for reaction-time analysis and manual validation of the automatic labels. EMBER reaches $73.5\%$ and ORBIT $71.2\%$ on the full validation set against a $51.5\%$ majority-class baseline; both generalize to held-out players. The strongest large language model baseline reaches $62.7\%$ on 1,000 predictions drawn from the same validation set; on those same trials EMBER ($73.6\%$) and ORBIT ($70.9\%$) outperform it (McNemar, $p = 1.3\times10^{-9}$ and $p = 1.5\times10^{-6}$), and the best LLM accuracy did not improve across a generation of newer models. Both models run at several orders of magnitude lower inference time and cost than the LLM baselines, although not compute-matched. A short post-anchor window turns prediction into early response recognition, with accuracy rising monotonically to $85.15\%$; we propose this curve as a measurable handle on the fast-to-deliberative control transition, a hypothesis that the present, roughly uniform per-tick gains support but do not establish. Neuroscience vocabulary is hypothesis-generating rather than mechanistic evidence; physiological ground truth, control distributions, and the reverse-inference constraint remain open, and a roadmap toward auditory reconstruction, larger-scale data, and telemetry-native foundation models closes the paper.
\end{abstract}


\section{Introduction}
Human behavior under pressure is not governed exclusively by slow, deliberative reasoning. Contemporary neurophysiology suggests that rapid, evolutionarily older systems can initiate orienting and action before full cortical appraisal is complete, and coarse sensory detection, salience evaluation, and motor preparation may occur within approximately 60--120~ms with limited conscious involvement. We treat this interval as an operational research window rather than a settled neuroscience category; its strongest direct support is intracranial evidence of rapid fear-selective amygdala responses in humans, approximately 74~ms for visible and 88~ms for invisible fearful faces~\cite{mendezbertolo2016,ywang2023}. The \mesocort{} project asks whether traces of such fast, salience-driven, limbic-like control can be detected and modeled from high-resolution Counter-Strike~2 (CS2) gameplay telemetry~\cite{pessoa2010,kahneman2003,mcfadyen2017}.

Dual-pathway theories distinguish a slower cortical ``high road'' from a faster subcortical ``low road,'' but the framing is debated: Pessoa and Adolphs argue for ``many roads'' rather than a single privileged shortcut, and recent discussion emphasizes distributed relevance evaluation over a one-route model~\cite{pessoa2010,ledoux1996,pessoa2013}. Converging intracranial, primate, and tractography evidence nonetheless keeps a fast subcortical contribution plausible~\cite{mendezbertolo2016,ywang2023,inagaki2023,rafal2025}. In motor neuroscience, the StartReact phenomenon shows that, under some conditions, prepared motor programs can be released at unusually short latencies through brainstem-mediated pathways; it serves here only as an analogy, since many CS2 responses are clearly more complex than a preloaded laboratory movement and should not be read as direct StartReact equivalents~\cite{vallssole1999,carlsen2009,adjei2025}. These perspectives motivate the hypothesis that at least some actions in high-pressure digital environments may emerge from anticipation, rapid salience confirmation, and fast motor execution rather than from fully deliberative reasoning alone.

CS2 combines millisecond-level reaction demands, incomplete information, repeated threat-like stimuli, and rich behavioral telemetry. It thus serves as a naturalistic laboratory that preserves the complexity of real decision-making under pressure at resolutions rarely available experimentally; unlike classical reaction-time experiments isolating simple sensorimotor responses, it captures the continuous interaction of reflexive reactions, expectation, tactical reasoning, and strategic adaptation~\cite{wang2023,aliyari2018,nobre2018}.

Replay data contains the full ground truth of the match, whereas the player does not; omniscient access to the world state makes standard AI systems poor models of human cognition. The project's core methodological principle is therefore an \emph{Ego-Knowledge} layer approximating what the player could realistically see, hear, infer, and believe at each moment.

On top of this layer, \emph{first-contact events}, visual or auditory moments anchoring time zero for reaction latency, support analysis of rapid post-stimulus behavior (crosshair correction, flick onset, counter-strafing, crouching, escape movement) as candidate indicators of a fast reactive layer, distinct from slower tactical actions (peeking, holding, retreating). This yields a hierarchy of behavioral targets (\textbf{H1}, micro-intent and reflexive motor response; \textbf{H2}, tactical choice; \textbf{H3}, strategic plan shifts), with the longer-term aim of inferring not only the action but the control mode behind it; the present paper addresses the action.

Modeling proceeds incrementally, from baseline models that validate feature quality and leakage risk toward sequential models over Ego-Knowledge history evaluated on unseen players, maps, and regions; if successful, this would offer a framework for studying pressure-sensitive behavior in ecologically valid digital environments~\cite{wang2023,murty2023}.

\textbf{Scope of the neuroscience.} The neuroscience reviewed in the next two sections is the motivation for this work, not one of its results. It shaped three design choices: which events we extract (abrupt, high-salience contacts), where we look (the first moments after stimulus onset), and the vocabulary of the outcome labels. The contributions and results themselves are behavioral and computational: a representation of what a player could perceive, a pipeline that extracts reaction moments, and models that predict a labeled outcome. No result in this paper localizes a response to a brain structure or identifies the control process that produced it, and the system latency of the recordings means that the 60--120~ms window cannot be resolved on single events (Section~\ref{sec:latency}). We use ``limbic-like'' descriptively, for fast salience-driven defensive responding, without presupposing a unitary limbic system, a concept that has long been questioned~\cite{ledoux2000}. Section~\ref{sec:limitations} states what the present data cannot show, and Section~\ref{sec:future-work} lists the experiments that would test the framing.

\subsection{Contribution Framing}

We also introduce two complementary \textit{fight-or-flight} classifiers: ORBIT, a custom transformer for temporally ordered Ego-Knowledge sequences, and EMBER, a gradient-boosted, non-sequential model for lower-cost prediction from aggregated game-state features. Both are evaluated against commercial and open-source large language model baselines, and we also analyze performance as the input extends into the post-anchor window, separating pre-reaction prediction from early response recognition. Although EMBER performs on par with or slightly above ORBIT in this constrained Phase~1 setting, the transformer is the primary direction because the longer-term objective is a single model serving multiple behavioral use cases under substantially larger, more temporally structured data.

The empirical anchor is a deliberately constrained class of high-salience back-damage events. Their stimulus-response structure is clear: the attack arrives from outside the current focus of attention, and the response reads as an immediate defensive, evasive, or engagement-oriented reaction, less ambiguous than complex tactical situations.

The primary contributions of this work are as follows:
\begin{enumerate}[leftmargin=*, itemsep=2pt]
    \item A conceptual framework for studying fast pressure-sensitive behavior in a high-resolution digital environment, with a modeling roadmap toward player-specific, multi-timescale behavior prediction by a single telemetry-native model serving multiple behavioral tasks at larger data scale.
    \item A leak-reduced Ego-Knowledge reconstruction pipeline for approximating player-specific subjective information and belief state.
    \item A custom millisecond-precision visibility engine that computes the visible fraction of a target from an observer's exact viewpoint via 3D depth-map segmentation, enabling a precise, retroactive stimulus-onset anchor not available from standard line-of-sight flags.
    \item A reaction-centered dataset focused on interpretable back-damage events, where the player is attacked from behind and must react immediately to an unexpected threat.
    \item A predictive modeling framework for \textit{fight/flight} outcome classification, combining coarse behavioral labeling with two complementary models, a custom transformer (ORBIT) and a gradient-boosted model (EMBER), and comparative evaluation against commercial and open-source LLM baselines.
    \item A manual gold-standard validation of the automatic labeling procedure (majority voting over seven expert annotators, with quantified consensus coverage and label agreement), and an initial empirical validation that back-damage events preserve a structured short-latency response regime in the source telemetry, with a reference-frame analysis reconciling the measurement floor, the proposed 60--120\,ms window, and the engine/network system delay separating demo-measured latency from true reaction time.
\end{enumerate}

\section{State of the Art}
Recent research suggests that virtual threat processing is not cognitively isolated from real-world threat processing: rather than a separate ``digital'' pathway, the nervous system appears to recruit conserved threat-related computations spanning sensory detection, integrative limbic and hypothalamic processing, and survival-oriented output structures such as the periaqueductal gray, shaped in games by amygdala and prefrontal interaction. In virtual reality, oscillatory correlates of threat imminence have been reported during first-person navigation, and early threat perception appears partially dissociable from later cognitive control~\cite{mcfadyen2017,murty2023,mobbs2009,karpov2024,lu2023}.

A useful organizing framework is the \emph{Predatory Imminence Continuum}, which distinguishes pre-encounter, post-encounter, and circa-strike defensive modes, from cautious scanning and anticipatory vigilance through freezing and risk assessment to explosive \textit{fight-or-flight} when threat becomes proximal and unavoidable; central-amygdala circuitry has been proposed as a candidate switching mechanism~\cite{mobbs2009,fanselow2018,fanselow2022,moscarello2022,abend2023}. For gameplay analytics this is an analogy rather than a literal equivalence (a player peeking a corner is not an animal in post-encounter defense), but it provides a useful computational vocabulary for high-pressure behavioral regimes, especially alongside the emerging VR evidence~\cite{karpov2024,abend2023}.

Digital games are nonetheless not perfect equivalents of real-world threat. Stressful 2D gaming environments produce measurable autonomic and psychological stress responses but also involve \emph{somatic decoupling}: threat-related arousal occurs while the downstream motor consequences of biological \textit{fight-or-flight} are constrained by the sedentary, interface-mediated task. \mesocort{} therefore does not claim direct measurement of literal biological \textit{fight-or-flight} behavior, but seeks \emph{limbic-like behavioral signatures} or \emph{low-latency salience-driven response patterns} that emerge under pressure in a high-resolution digital setting~\cite{wang2023,wells2022,aliyari2018}.

In-game avoidance, withdrawal, aggression, pacing, and risk-taking may reflect broader tendencies such as coping style, neuroticism, or behavioral inhibition, although fast behavior in highly practiced players may instead reflect automaticity and overlearned sensorimotor routines~\cite{laier2018,fitts1967}; the defensible implication is not personality inference from isolated episodes but that repeated player behavior may contain stable personal signatures.

fMRI studies of practiced FPS players report active suppression or deactivation of amygdala responses during gameplay, with increased engagement of dorsal anterior cingulate and prefrontal regions associated with executive control, suggesting that with extensive practice threat processing becomes increasingly cortically regulated rather than subcortically driven~\cite{montag2012,weber2006,klasen2012}. The population studied here, experienced competitive players with thousands of hours of play, may therefore be the population \emph{least} likely to exhibit tonic limbic engagement during routine gameplay. This does not invalidate the search for fast salience-driven responses, since tonic suppression is compatible with transient phasic reactivation during genuine surprise events such as unexpected back-shots, but any ``limbic-like'' signal in experienced players is likely to be event-locked, brief, and substantially attenuated relative to naive subjects. We treat this literature as a boundary condition rather than a contradiction: it narrows the expected signal to specific high-salience surprise events and argues against reading broad gameplay behavior as limbic.

Repeated exposure to digital stressors may also lead to habituation or desensitization, and the absence of real biological cost changes risk calibration relative to physical reality, arguing against strong claims of one-to-one translation between virtual and physical behavior. The most robust framing is that game telemetry offers an ecologically rich environment for studying \emph{pressure-sensitive behavioral organization}, while broader transfer claims must remain probabilistic, constrained, and empirically validated~\cite{pessoa2013,wang2023,wells2022}.

A second tradition bears directly on the control-mode question. Computational accounts of decision making distinguish model-free control, which selects actions from cached values learned through experience, from model-based control, which evaluates actions against an internal model of the environment, and treat behavior as arbitrating between them according to their relative uncertainty~\cite{daw2005,daw2011}. Two properties of that literature constrain the present work. First, model-free control is itself predictive: a temporal-difference value function is a prediction, so speed does not imply the absence of prediction. Second, model-based control becomes fast and automatic with training, and intermediate representations such as the successor representation can produce model-based-looking behavior through model-free machinery at low computational cost~\cite{russek2017}. Latency therefore cannot by itself identify which controller produced a response. We consequently name the reaction-time bands in this paper by latency rather than by mechanism, and treat the mapping from band to control mode as a hypothesis the design is built to test rather than as an assumption it encodes.

Relatedly, work on large-scale human gameplay data has established the practice of estimating an upper bound on predictable behavior by fitting a high-capacity neural network and treating its performance as a ceiling against which interpretable cognitive models are compared~\cite{kuperwajs2023}. We do not yet report such a bound. Section~\ref{sec:label_validation_majority_voting} reports only the reliability of our labels, whose Fleiss's kappa of $0.38$ indicates that a substantial part of the residual error may be attributable to label noise rather than to unmodeled behavioral structure. We identify the predictability ceiling as the primary missing quantity for interpreting the accuracies reported here.

The behavioral signal studied here is a high-frequency multivariate time series: each tick contributes a fixed-width feature vector, and the task maps a window of recent ticks to near-future behavior. Transformer forecasters such as Informer~\cite{zhou2021informer}, Autoformer~\cite{wu2021autoformer}, and PatchTST~\cite{nie2023patchtst} are strong on long-horizon benchmarks, but the quadratic cost of self-attention and its weak inductive bias for ordered numerical data have drawn criticism, most pointedly the finding that a simple linear model can outperform them~\cite{zeng2023ltsflinear}. State-space models, which scale linearly with sequence length but compress history into a fixed-size state and must retrofit cross-channel interaction~\cite{zeng2024cmamba,liu2024dpem,kang2025dcmamber,chen2026dmamba,wang2024smamba}, and pretrained time-series foundation models~\cite{ansari2024chronos,woo2024moirai,ekambaram2024ttm} are natural later directions. Because the relevant context spans hundreds of densely sampled ticks and the per-tick feature vector is strongly multivariate, we adopt a custom attention-based architecture (Section~\ref{sec:transformer}).

\section{Neurophysiological Motivation for the Fast Reaction Window}
\mesocort{} hypothesizes a fast reaction window of approximately 60--120~ms, treated as an operational research interval rather than a canonical neuroscience construct. The lower end is anchored by intracranial evidence of fear-selective amygdala responses at approximately 74~ms for visible and approximately 88~ms for invisible fearful faces in humans~\cite{mendezbertolo2016,ywang2023}; the upper end loosely marks where cortically elaborated processing becomes increasingly plausible. It is thus not a reflex interval but a zone in which rapid salience detection, anticipatory motor readiness, and fast descending control may converge~\cite{pessoa2010,mcfadyen2017}.

One basis is the dual-pathway account of perception, described as a slower cortical ``high road'' and a faster, coarser ``low road.'' Conscious recognition depends on the geniculo-striate pathway (lateral geniculate nucleus, primary visual cortex), whereas a faster subcortical route via the superior colliculus, pulvinar, and amygdala has been proposed to detect coarse, potentially threatening patterns fast enough to bias behavior before detailed analysis concludes. This account remains debated, and the paper treats the ``low road'' as one plausible mechanistic contributor rather than as settled doctrine~\cite{mendezbertolo2016,ywang2023,pessoa2010,ledoux1996,inagaki2023,rafal2025}.

The amygdala is thus best understood as an early relevance-evaluation node rather than a passive recipient of interpreted percepts, and its rapid responses are consistent with threatening cues influencing defensive preparation at very short latencies. Some of the earliest post-contact adjustments in gameplay may therefore reflect fast salience-driven processing rather than only deliberate tactical reasoning~\cite{ywang2023,mcfadyen2017,inagaki2023}.

Fast response also depends on motor release: the corticospinal tract is more associated with fine, fractionated voluntary control, whereas the reticulospinal system is primarily associated with posture, coordination, and rapid whole-body action (with some fine-motor contributions)~\cite{riddle2010}. The \emph{StartReact} literature suggests that prepared motor programs can be released at unusually short latencies through brainstem-mediated mechanisms, especially under anticipatory readiness~\cite{vallssole1999,carlsen2009,adjei2025}. Here this is only a partial analogy: a CS2 player reacting to a back-shot is not necessarily executing a prepared motor program and may require directional inference and action selection.

Individuated finger movements are not susceptible to StartReact, being predominantly corticospinally mediated~\cite{carlsen2009}, but CS2 play involves bimanual coordinated control closer to the coordinated-grasp condition in which StartReact \emph{is} triggered~\cite{honeycutt2013}, and mouse aiming engages proximal musculature with well-established reticulospinal contributions~\cite{santangelo2023}. CS2 motor responses are therefore best described not as reticulospinally dominated but as a mixture of corticospinal and reticulospinal contributions whose weighting likely depends on the action (fine crosshair adjustment versus gross emergency reorientation) and on anticipatory preparation.

Very short latencies are therefore not automatically stimulus-driven. Between roughly 40 and 60~ms lies a gray zone in which sensory registration may begin but conscious discrimination and large-scale motor implementation are not yet plausible, so responses that appear ``too fast'' likely involve a mixture of prediction, anticipation, motor preloading, or rapid confirmation. Predictive coding reinforces this: in practiced, temporally structured environments such as esports, expectations may pre-activate sensorimotor pathways and reduce effective latency once sensory confirmation arrives, so exceptionally fast actions in Counter-Strike~2 should not be interpreted as ``pure reflexes'' but as the product of a coupled system of anticipatory preparation, rapid salience detection, and optimized motor execution~\cite{nobre2018,grabenhorst2019}.

For \mesocort{}, low-latency events after first contact may thus carry information about the player's control mode, and the 60--120~ms regime is especially interesting as the boundary between deliberate and pre-deliberative control. This justifies no claim of direct measurement of a brain structure or literal limbic discharge, but it does justify treating them as candidate markers of a fast reactive layer of behavior: neurophysiologically plausible, computationally measurable, and distinct from slower tactical and strategic planning~\cite{mcfadyen2017,carlsen2009,nobre2018}. In the present data the window itself is not resolvable: the recordings carry a system delay of the same order as the window (Section~\ref{sec:latency}), so it serves here as the motivation for the latency analysis rather than as a quantity we measure.

\section{System Overview and Ego-Knowledge Reconstruction}
\begin{figure*}[htp]
    \centering
    \includegraphics[width=\textwidth]{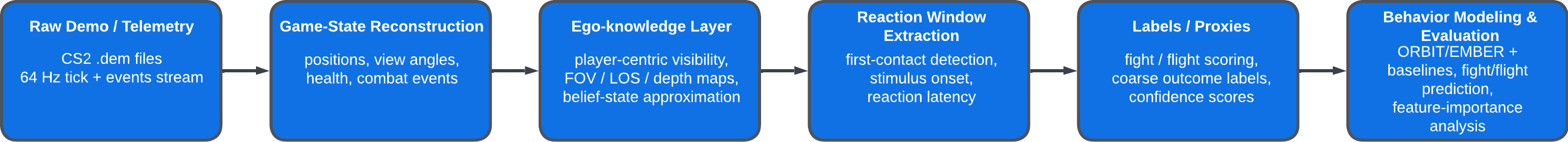}
    \caption{High-level architecture of the Mesocort pipeline. Raw CS2 demo telemetry is transformed into reconstructed game state, player-centric Ego-Knowledge, reaction windows, and fight/flight labels used for behavior modeling, prediction, and evaluation.}
    \label{fig:pipeline}
\end{figure*}

The system has four layers (Fig.~\ref{fig:pipeline}): capture acquires gameplay traces and game state from CS2 demo files, and reconstruction transforms them into a player-centric representation of what could plausibly be perceived at each moment. Event mapping extracts first-contact windows and downstream actions, and prediction uses the resulting sequence for short-horizon behavior modeling.

\subsection{Why Ego-Knowledge Is Necessary}
A replay contains the full ground truth; the player sees only a restricted, uncertain subset shaped by line of sight, field of view, sound cues, prior observations, map knowledge, and short-term inference. A model trained on omniscient replay truth may predict well yet remain cognitively unrealistic. \mesocort{} therefore centers on an Ego-Knowledge layer approximating the subjective information state, not the objective replay truth.

This distinguishes \mesocort{} from environment world models such as JEPA and V-JEPA~2~\cite{lecun2022,vjepa2_2025}, which learn environment dynamics but do not model the gap between the objective world state and the partial information available to a specific human agent; a world model trained on CS2 replays would risk omniscient leakage. Ego-Knowledge does not reconstruct the full world state but filters it through what the player could plausibly perceive or infer (Section~\ref{sec:discussion-conceptual}).

\subsection{Reconstruction Principles}
A leak-reduced pipeline converts demo files into a per-tick timeline of player positions, view angles, utility usage, health states, and combat events. Line-of-sight and field-of-view calculations estimate what was visually available at each moment; auditory estimation is planned, fusing both into a belief state of where the player could plausibly believe enemies to be. This remains a physically motivated approximation rather than a validated proxy for full perception: it does not quantify attentional narrowing, display-dependent detectability, or other observer-side factors, and reconstruction fidelity is an open validation target. The current layer is hand-engineered; the target architecture is a learned agent-centric encoder, analogous to I-JEPA~\cite{lecun2022}, with the visibility engine as a training-time data filter, not a feature extractor.

\subsection{Behavioral Hierarchy}
\textbf{H1} captures micro-intent and reflexive motor response: abrupt crosshair correction, movement onset, weapon fire, or rapid reorientation after first contact. \textbf{H2} captures tactical choices: peeking, holding, retreating, rotating, or deploying utility. \textbf{H3} captures strategic plan shifts: delayed team rotation or map-wide repositioning. A sub-100~ms motor onset and a multi-second strategic response do not share a control process. The mapping to threat-imminence theory is analogical, not literal, but bridges established neuroscience frameworks~\cite{moscarello2022,abend2023}: pre-encounter states (rotating to a new area, no enemies detected, high uncertainty) map to strategic control (H3); post-encounter states (enemy detected or suspected via sound, communication, or partial visibility) to tactical control (H2); and circa-strike states (direct engagement, being shot at, surprise contact) to reactive control (H1).

\subsection{Definition of Fight/Flight Labels}
\textit{Fight} and \textit{flight} are operational gameplay labels, not direct references to the classical \textit{fight-or-flight} response described by Cannon~\cite{cannon1915bodily}: a coordinated acute stress reaction in which sympathetic and adrenomedullary activation mobilize the organism for defensive action~\cite{cannon1915bodily,godoy2018stress,purves2001neuroscience}. Our use is narrower and purely behavioral: the labels imply neither a literal survival response nor direct observation of sympathetic, hormonal, limbic, or other neurophysiological mechanisms, only their measurable outcomes in high-frequency gameplay data.

Labels are defined around reaction moments in which the analyzed player is damaged from outside the current field of view, most characteristically from behind, a scenario chosen for its relatively clear stimulus-response structure: a sudden, unexpected threat answered under time pressure with limited information on the attacker's position.

A reaction is labeled \textit{fight} when immediate post-damage behavior indicates re-engagement with the threat (rapid rotation toward the attacker, crosshair correction, visual search, aiming, or fire toward the source of damage) and \textit{flight} when it indicates disengagement (movement toward cover, retreat from the exposed position, repositioning away from the attacker, or evasive movement without an immediate attempt to locate or engage the opponent). These are the dominant observable alternatives after the triggering event: coarse outcome classes summarizing the short-horizon response, not exhaustive descriptions of intention. \textit{Fight} often contains H1 components; \textit{flight} often adds H2-level repositioning.

\section{Data and Data Preparation}
\label{sec:dataset}

The data foundation is a medallion architecture with three layers, stored as Parquet-backed Delta Lake tables: \emph{Bronze} preserves raw telemetry in immutable form, \emph{Silver} consolidates it for analysis, and \emph{Gold} prepares model-ready features. The atomic Bronze unit is a \emph{tick}, one sampled snapshot of the complete game state; CS2 demos are recorded at 64~ticks per second, approximately $15.625$~ms per sample. Bronze separates into \emph{tick-level data}, a per-tick state vector for every player (position, velocity, team, health, armor, inventory, view orientation, pressed buttons, and round-level context), and \emph{event-level data}, a sparse stream of discrete combat, bomb-lifecycle, and meta events. Silver removes invalid or incomplete matches, applies consistency checks against noise and leakage, adds match summaries, and derives features aligning sparse events with the continuous game state. Gold produces the wide feature vectors consumed by training: numerical features are scaled according to their distribution (min-max, standard, power, and logarithmic transformations) and categorical and discrete spatial variables are one-hot encoded or indexed for embedding layers.

\subsection{Raw Data Source: CS2 Demo Files}
CS2 stores every match as a binary demo file, a compact tick-by-tick log of the game state, which we decode with a custom high-performance C++ parser into the per-tick data used here. All data come from server-side GOTV demos, which record all players from the server's authoritative view, the source of the server-versus-client timing gap discussed in Section~\ref{sec:latency}.

\subsection{Dataset features}

Each input vector describes the reconstructed game state at a single tick from the analyzed player's perspective; the model receives a sequence of $N$ consecutive such vectors. Features fall into seven semantic groups (Table~\ref{tab:feature_group_summary}). Player features describe position, view direction, health, equipment, movement, and objective-related state; positions of the player, teammates, and enemies are represented both as continuous coordinates and as discretized map-square identifiers used as embedding indices. Aiming and visibility features describe crosshair movement, angular and Euclidean distances to enemies, enemy visibility, and recent observation history, some of which summarize short-term history (e.g., ticks since an enemy was last visible). Player input features encode button states and weapon firing; damage and kill features summarize recent combat interactions and their temporal distance from the current tick. Teammate features cover distance, visibility, and position; enemy features add alive state, line-of-sight relation, crouching, firing, visibility ratio, crosshair distance, and radial velocity. In total, each tick is represented by 154 input features; the complete list with preprocessing methods is given in Appendix~\ref{sec:appendix-input-features}, Table~\ref{tab:dataset_features_all}.

\begin{table*}[t]
\centering
\caption{Summary of input feature groups used by the model.}
\label{tab:feature_group_summary}
\begin{tabularx}{\textwidth}{l c X}
\toprule
\textbf{Feature group} & \textbf{No.} & \textbf{Description} \\
\midrule
Round and map context & 5 & Round-level and map-level state variables. \\
Player position and state & 21 & Player position, view direction, equipment, movement, and status variables. \\
Aiming and visibility & 13 & Crosshair movement, enemy visibility, and enemy proximity features. \\
Player input actions & 12 & Button states and weapon-firing indicators. \\
Damage and kill events & 14 & Tick-level and round-level damage and kill history features. \\
Teammate features & 24 & Position and visibility features for four teammates. \\
Enemy features & 65 & State, position, visibility, distance, and movement features for five enemies. \\
\midrule
\textbf{Total} & \textbf{154} & All input features computed for the current game tick. \\
\bottomrule
\end{tabularx}
\end{table*}

\subsection{Source Domain and Sampling Strategy}
For Phase~1 of \mesocort{} modeling, we constructed a reaction-centered dataset from Counter-Strike~2 demo recordings, selecting 50 players with the most stable game history and collecting their most recent demos from approximately the last 6 months: an initial pool of approximately 35{,}000 games, about 700 per player. The rank composition was a compromise between partial uniformity and the natural rank distribution of the broader game population.

\subsection{Reaction-Moment Scenario}
The dataset is built around the \emph{reaction moment}: a temporally localized situation in which a player must respond to a salient, combat-relevant stimulus under time pressure. The prototypical case is a one-on-one confrontation in which the victim sustains damage from an attacker outside the current field of view, most characteristically a shot from behind. This imposes a clean stimulus-response structure: the threat arrives abruptly, the player is not visually prepared, and the immediate behavior can be interpreted as a defensive, evasive, or retaliatory reaction to a single well-defined onset, with minimal confounding from teammates, crossfire, and competing threats. In Phase~1, candidate moments were triggered by an enemy becoming visible or by the player sustaining damage; prior internal analysis indicated similar reaction-time distributions across the two, so both were included in a shared dataset. The damage-from-behind subset is treated as the cleanest interpretable case and anchors the reaction-time analysis.

\subsection{Leak Reduction and Filtering Logic}
A central design goal was to reduce hidden confounds that could distort the interpretation of reaction latency. Because a full auditory engine was not yet implemented in Phase~1, the extraction procedure explicitly attempted to mitigate sound-related ambiguity. Reactions to thrown utility were excluded, and later iterations added rules removing cases such as flashed players or active shootouts, which proved to heavily influence the player's ability to react. These constraints keep the extracted moments interpretable with respect to immediate visual or direct-damage triggers. The network and demo-recording delays that shift the effective stimulus-onset anchor are discussed in Section~\ref{sec:latency}.

\subsection{Dataset Composition and Splits}
For the present experiments we intentionally use only a selected subset of our broader data estate: Phase~1 prioritizes interpretability, leakage reduction, and controlled validation over scale, and the subset is a research sample drawn from a substantially larger and continuously growing platform.\footnote{The broader infrastructure has processed more than 4.5 million \texttt{.dem} files, covering over 1{,}200~TB of gameplay data and approximately 3{,}000 years of cumulative game time; these figures are context, not the dataset analyzed here.}

As summarized in Table~\ref{tab:dataset_summary}, the experimental corpus comprises 32{,}205 unique demos, 27{,}356 for training and 4{,}849 for validation. With the filters above, extraction isolated 271{,}541 reaction moments, approximately 8 per demo, across 119{,}149 unique players; the 50-player core subset contained 28{,}332 moments, roughly 567 per player. Triggers were strongly dominated by visual exposure: 251{,}427 moments (92.6\%) were triggered by an enemy becoming visible and 20{,}114 (7.4\%) by damage received. A semi-automated rule-based procedure then assigned a coarse outcome of \emph{fight} (the decision to engage an opponent in a duel) or \emph{flight} (the decision to retreat) from relevant features in the 2~seconds following the trigger. In the current version of the dataset, 131{,}628 reactions (48.5\%) are labeled \textit{fight} and 139{,}913 (51.5\%) \textit{flight}; the pipeline also stores a confidence-oriented score indicating how strongly a reaction supports its assigned class.

\begin{table}[t]
\caption{Phase~1 reaction dataset summary.}
\label{tab:dataset_summary}
\centering
\begin{tabular}{p{0.53\columnwidth}p{0.34\columnwidth}}
\toprule
\textbf{Statistic} & \textbf{Value} \\
\midrule
Core players selected for Phase~1 & 50 \\
Average demos collected per core player & 690 \\
Initial pool of matches & $\sim$35{,}000 \\
Time span covered up to & $\sim$7 months \\
Filtered demos in extracted dataset & 32{,}205 \\
Total reaction moments & 271{,}541 \\
Average reaction moments per demo with filters & $\sim$8 \\
Unique players observed overall & 119{,}149 \\
Reaction moments in 50-player core subset & 28{,}332 \\
Average reaction moments per core player & $\sim$567 \\
Visibility-triggered moments & 251{,}427 (92.6\%) \\
Damage-triggered moments & 20{,}114 (7.4\%) \\
Fight labeled reactions & 131{,}628 (48.5\%) \\
Flight labeled reactions & 139{,}913 (51.5\%) \\
\bottomrule
\end{tabular}
\end{table}

\subsection{Custom Visibility Engine for Millisecond-Precision Analysis}
\label{sec:visibility-engine}
\begin{figure}[h]
    \centering
    \includegraphics[width=1.0\linewidth]{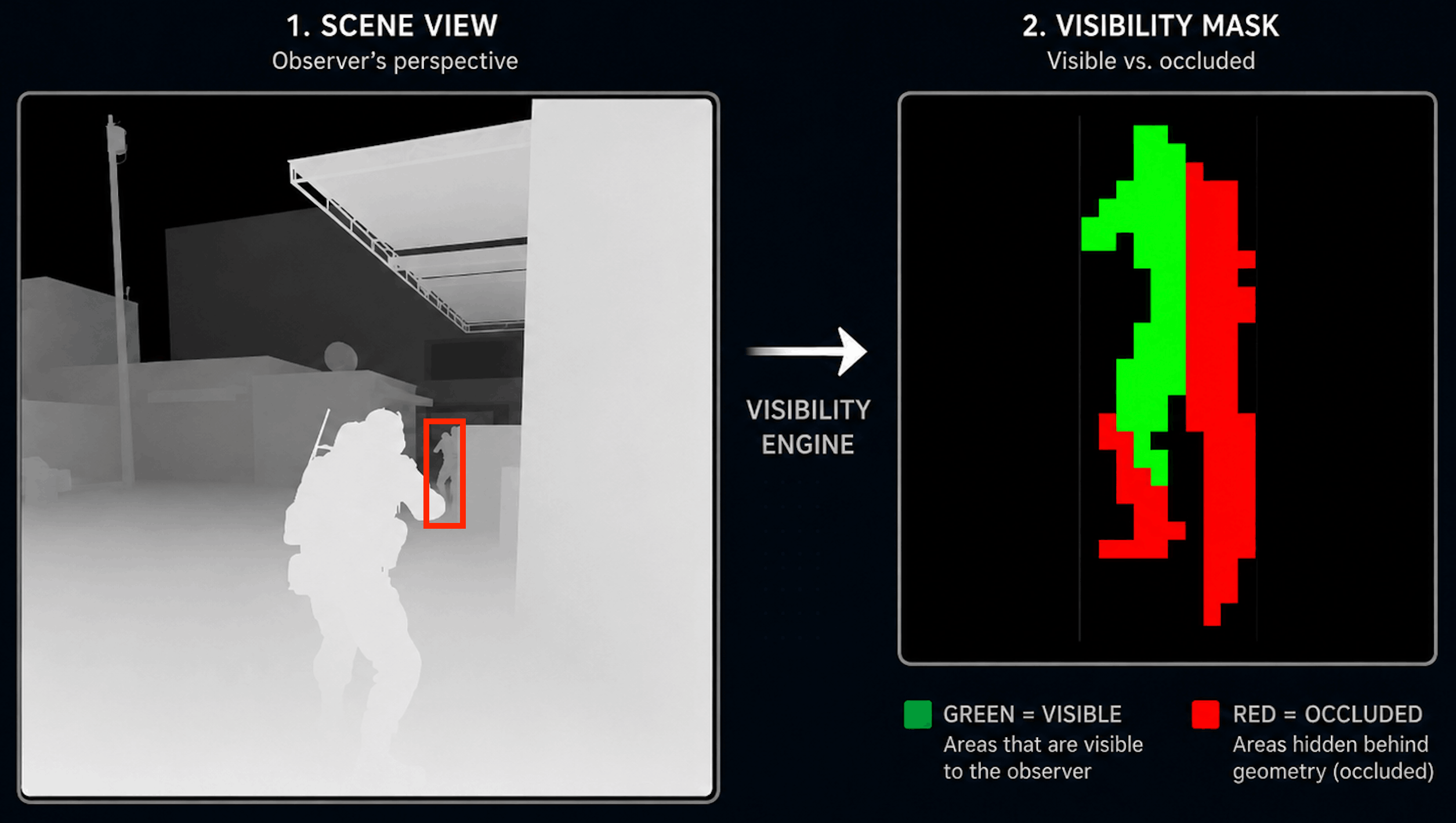}
    \caption{Visibility engine processing flow: on the left, the scene's depth map showing the silhouette of the target player; on the right, the pixel segmentation result, where green marks the visible portion and red the portion obstructed by geometry (e.g., a wall), from which the per-frame visibility percentage is computed.}
    \label{fig:visiblity_engine_2}
\end{figure}

For analysis at the scale of fractions of a second, we developed a custom visibility engine, since the required signal is not natively available in standard game outputs. It computes the visible fraction of an enemy model from the observer's point of view for every frame using 3D depth-map simulations of the full game geometry: for each tick, a depth map is rendered from the observer's exact viewpoint, the target player's model is rendered into the same scene as a silhouette mask, and pixel-level segmentation marks each silhouette pixel visible (green in Fig.~\ref{fig:visiblity_engine_2}) where the model is closer than the nearest occluding surface, or obstructed (red) where geometry intervenes. The ratio of visible to total silhouette pixels gives a per-frame visibility percentage, which identifies the exact moment visual detection begins. For visibility-triggered moments this percentage defines the stimulus-onset anchor from which latency is measured, so any imprecision propagates directly into the behavioral analysis; for damage-triggered moments such as back-damage events, the onset is anchored to the damage tick. Running retroactively on recorded demos, it applies at scale without real-time rendering. To our knowledge no comparable system exists in the public esports analytics literature, where visibility is typically approximated by coarse line-of-sight checks or server-side flags lacking the spatial resolution needed for sub-100~ms analysis.

\subsection{Reaction-Time Distributions}
Latency between stimulus onset and reaction onset is measured by an onset-detection algorithm that considers three observed signals (movement on the map, rotation of the player, and shooting) and searches their differences for a sudden change of predefined magnitude; if no plausible candidate is found, the situation is dropped. Reactions cannot be paired with concrete stimuli in 100\% of cases, but steps were taken to minimize the influence of other stimuli; the resulting error was found statistically insignificant on a randomly chosen subset and without emphatic bias. Manual labels on a subset of the same scenario moments seem likely to follow a similar distribution to the automatic ones, the most notable difference being more low reaction times (below 150~ms) in the automatic source. Distributions also vary among annotators, so reaction time is not unambiguous from the game data alone, although within a larger dataset these uncertainties are statistically diminished. The back-damage reaction-time histogram is reported in Section~\ref{sec:rt-results}.

\section{Proposed Methodology}

\begin{figure*}[t]
    \centering
    \includegraphics[width=\textwidth]{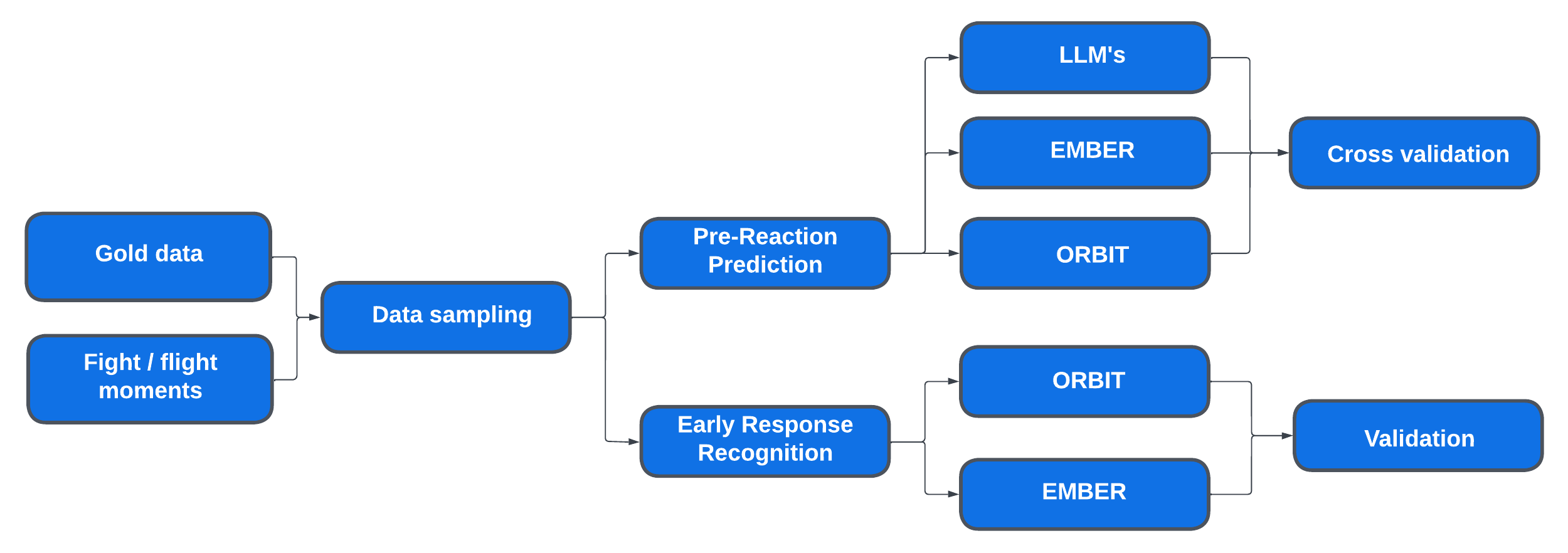}
    \caption{Overview of the methodology. Pre-reaction prediction uses only information up to the anchor moment; early response recognition may also use a short post-anchor window.}
    \label{fig:methodology_pipeline}
\end{figure*}

Figure~\ref{fig:methodology_pipeline} summarizes the methodology. Gold-layer data and extracted \textit{fight-or-flight} moments are combined into model-specific inputs; two settings follow from the temporal information given to the model. In \emph{Pre-Reaction Prediction}, models classify the player's later response from information available at or before the anchor (trigger) moment; LLM baselines, EMBER, and ORBIT are evaluated here. In \emph{Early Response Recognition}, ORBIT and EMBER receive windows that may include a short post-anchor segment, observing part of the reaction as it unfolds, with results reported on a validation split.

\subsection{Fight/Flight Label Generation}
\label{sec:labeling}

\subsubsection{Automatic Labeling}

The dataset contains first-contact events that led to a reactionary decision. A proprietary rule-based process labels the post-stimulus reaction by estimating the player's intention: to engage the threat (\emph{fight}) or to retreat (\emph{flight}).

The process considers player-specific signals expected to reflect intention in a two-second window after the reaction tick. For each signal and several sub-intervals of that window, the change in the signal is independently mapped to a \textit{fight-or-flight} score via quantile-based thresholds. Sub-intervals closer to the reaction carry higher weight, but the window extends to two seconds because a player can start reacting while the animation of a previous action is still finishing. The scores are combined into a weighted aggregate
\[
score = \sum_{s \in S} \sum_{t \in T} w_s w_t \, x_{s,t},
\]
where $S$ is the set of signals and $T$ the set of intervals; $w_{t}$ decreases as $|t|$ increases, $w_{s}$ is set from experience, and both the weights and the set of signals were re-assessed manually on a data subset. The final score is the absolute difference between the \textit{fight} and \textit{flight} scores, and the label is the one with the higher score.

The scoring system is thus a deterministic classifier of future player action; higher scores express higher confidence, and low-confidence values were removed from the dataset used here.

\subsubsection{Manual Labeling}
\label{sec:manual_labeling}
To obtain ground-truth labels for the response of an unexpectedly damaged player, we developed a custom annotation tool for short, isolated one-versus-one situations in which a player received damage from behind without initially seeing the attacker. A total of 200 such situations were selected, each centered on the damage event. The annotation provided a ground-truth reference for the automatic labels and identified inherently ambiguous situations.

Seven annotators with advanced knowledge of CS2 gameplay labeled the situations independently as \textit{fight} (the player took the duel: turning toward the attacker, aiming, or firing), \textit{flight} (the player disengaged: moving away, repositioning, or taking cover), or \textit{don't know} when the evidence was insufficient. Discrepancies were resolved by majority voting.

\begin{figure}[t]
    \centering
    \includegraphics[width=\linewidth]{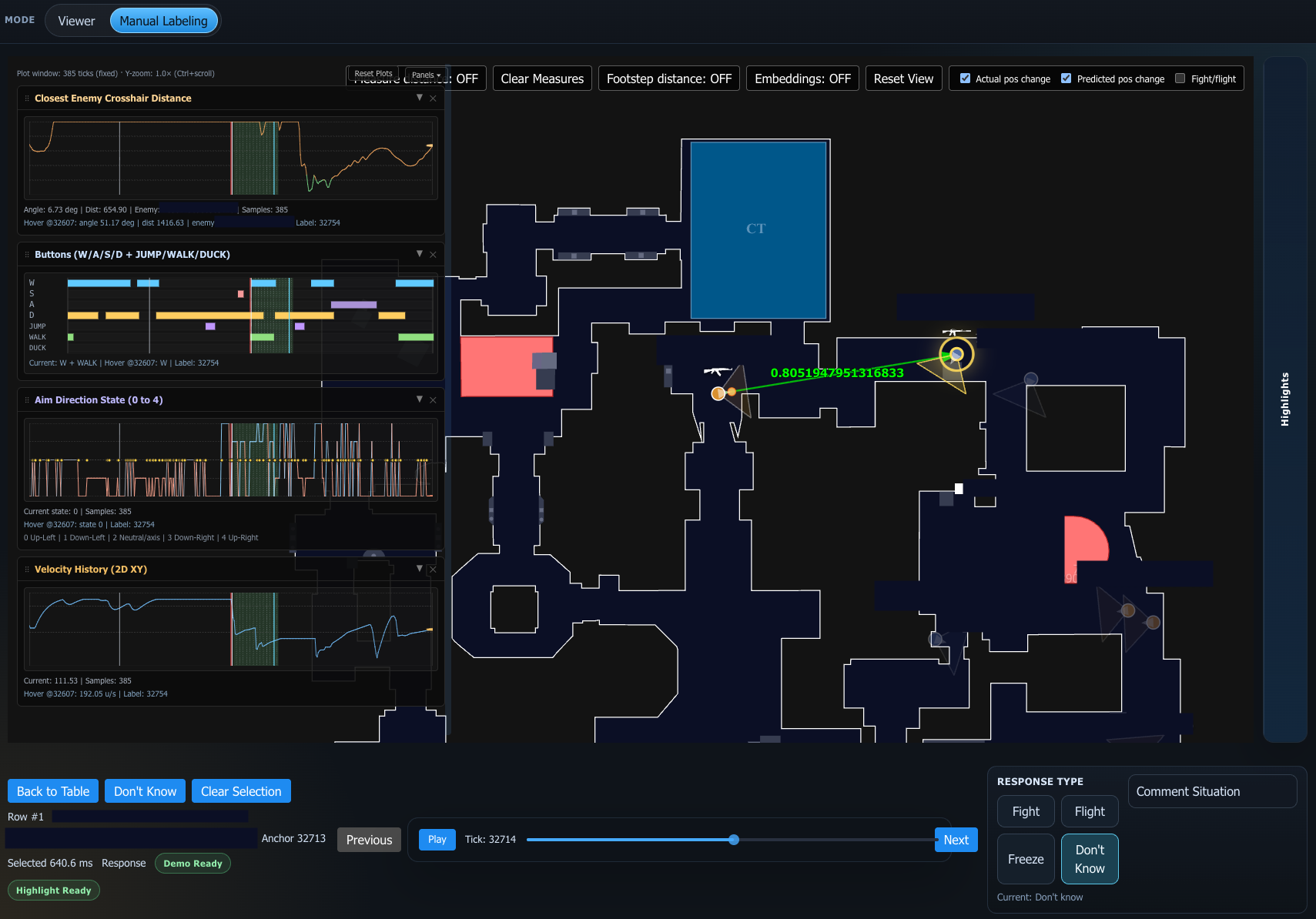}
    \caption{Manual annotation tool. The central view reconstructs the situation in two dimensions; the left panel shows synchronized temporal signals of the analyzed player; the remaining panels hold playback, annotation controls, and gameplay highlights.}
    \label{fig:manual_labeling_tool}
\end{figure}

The interface (Fig.~\ref{fig:manual_labeling_tool}) combines a two-dimensional top-down reconstruction of the game state with synchronized temporal signals of the analyzed player (crosshair movement, pressed buttons, aim-direction changes, velocity), playback and annotation controls, and gameplay highlights for visual confirmation.

\subsection{Model Task and Input Sampling}
\label{sec:model-taks-input-sampling}

Gold-layer data are chronologically ordered tick-level feature vectors, converted into model-specific samples. The task is defined around selected moments rather than all ticks: each labeled situation is centered on an \emph{anchor tick}, the moment at which the situation is identified, and the label describes the player's subsequent response as \textit{fight} or \textit{flight}, which typically occurs within approximately one second after the anchor.

Let $\mathbf{x}_{t}$ denote the feature vector at tick $t$ and $t_a$ the anchor tick. A general temporal sample is
\[
    \mathbf{X}_{t_a,H,F} =
    \left[
        \mathbf{x}_{t_a-H+1},
        \ldots,
        \mathbf{x}_{t_a},
        \ldots,
        \mathbf{x}_{t_a+F}
    \right],
\]
where $H$ is the number of historical ticks up to and including the anchor tick and $F$ the number of future ticks after it; the sample's label is the response associated with the anchor situation. A single-tick sample has $H=1$, $F=0$; a history-only sequence has $H=N$, $F=0$.

\subsubsection{Pre-Reaction Prediction}
Here models classify the response from information available at or before the anchor only ($F=0$): the post-anchor reaction is not observed, and the eventual label must be inferred from the current or preceding game state. For non-sequential models such as XGBoost the input is the anchor-tick vector $\mathbf{x}_{t_a}$ or, as in EMBER, a short window of subsampled pre-anchor snapshots (Section~\ref{subsec:EMBER}); for sequence models such as transformers it is a history window of $N$ ticks ending at the anchor, $\mathbf{X}_{t_a,N,0}$. The target is thus consistent across EMBER and ORBIT, while ORBIT can use ordered pre-anchor context.

\subsubsection{Early Response Recognition}
\label{subsec:early-response-recognition}
The second setting studies how post-anchor information ($F>0$) influences performance, with future windows restricted to the short reaction interval after the anchor, up to approximately one second (64 ticks). Although automatic labeling uses a longer two-second window to assign a robust outcome, the 64-tick window tests how early the response can be recognized from its initial development, keeping the focus on fast pressure-driven behavior. Because the model sees partial evidence of the actual reaction, results are interpreted as early response recognition rather than prediction from pre-reaction information; the label definition is unchanged and only the temporal extent of the input differs.

\subsection{Reaction Detection and Onset Extraction}
\label{sec:reaction-extraction}
Two time points must be fixed before a reaction moment is analyzable: the \emph{stimulus onset}, at which the triggering event becomes available to the player, and the \emph{reaction onset}, at which the player's behavior first reflects a response; moments in which either is ambiguous are rejected.

For visibility-triggered moments the stimulus onset is the first frame at which the enemy becomes visible (Section~\ref{sec:visibility-engine}); for back-damage events, where the attacker is typically outside the field of view, it is the damage tick, and a large yaw mismatch between view and attacker direction is additionally required so that the stimulus originates outside the current focus of attention.

Reaction onset is the first sudden change in player movement after the stimulus, operationalized for back-shots primarily as a rapid yaw change, assuming the canonical response is a turn toward the threat. It is localized from crosshair kinematics: the reaction begins when the player abruptly decelerates an ongoing movement before re-accelerating toward the threat, visible in the first and second derivatives of mouse speed. This deceleration and reacceleration signature separates stimulus-independent motion from directed corrective movement, although motion already underway from a previous event can blur it.

Confounded moments (the player already in a shootout, engaging another enemy, flashed, or otherwise occupied) are excluded, and latencies are searched between the minimum measurable latency ($\approx 46.875$~ms) and approximately one second. The lower bound excludes physically implausible or annotation-induced near-zero values but left-censors the distribution: responses faster than the measurable boundary accumulate near it and cannot, on their own, establish a distinct sub-100~ms mechanism. The extracted latencies are therefore descriptive rather than a formal decomposition of reaction-time regimes~\cite{ratcliff1979,ratcliff2008,luce1986}.

These latencies are measured in demo time, not true reaction time: $t_\text{measured} \approx t_\text{true} + \Delta_\text{stim} + \Delta_\text{resp}$, where the two strictly positive system delays total roughly $40$--$80$~ms for back-damage events. Measured onsets are thus conservative upper bounds, and only outcomes from approximately $100$~ms of measured latency onward are treated as candidate genuine reactions, the earliest range consistent with a true post-stimulus response once the delay is subtracted (Section~\ref{sec:latency}).

\subsection{Behavioral Prediction Model (ORBIT)}
\label{sec:transformer}
ORBIT is a transformer-based neural network that classifies a player's behavioral response from temporally ordered game-state representations. For a selected anchor tick, its input is a history window of $N$ consecutive tick-level feature vectors ending at the classified moment; $N$ is a configurable sequence length varied in the experiments.

Each tick-level feature vector becomes one input token holding the complete model input for the analyzed player at that tick (player state, spatial information, visibility, aiming-related features, input actions, and event-derived variables). Continuous and Boolean variables enter as numerical inputs; selected categorical or discrete variables (round number, weapon identifier, aim-direction state, discretized map-square identifiers) pass through embedding layers, and both are concatenated and projected into a fixed-dimensional token.

\begin{figure}[htbp]
    \centering
    \includegraphics[width=1.0\linewidth]{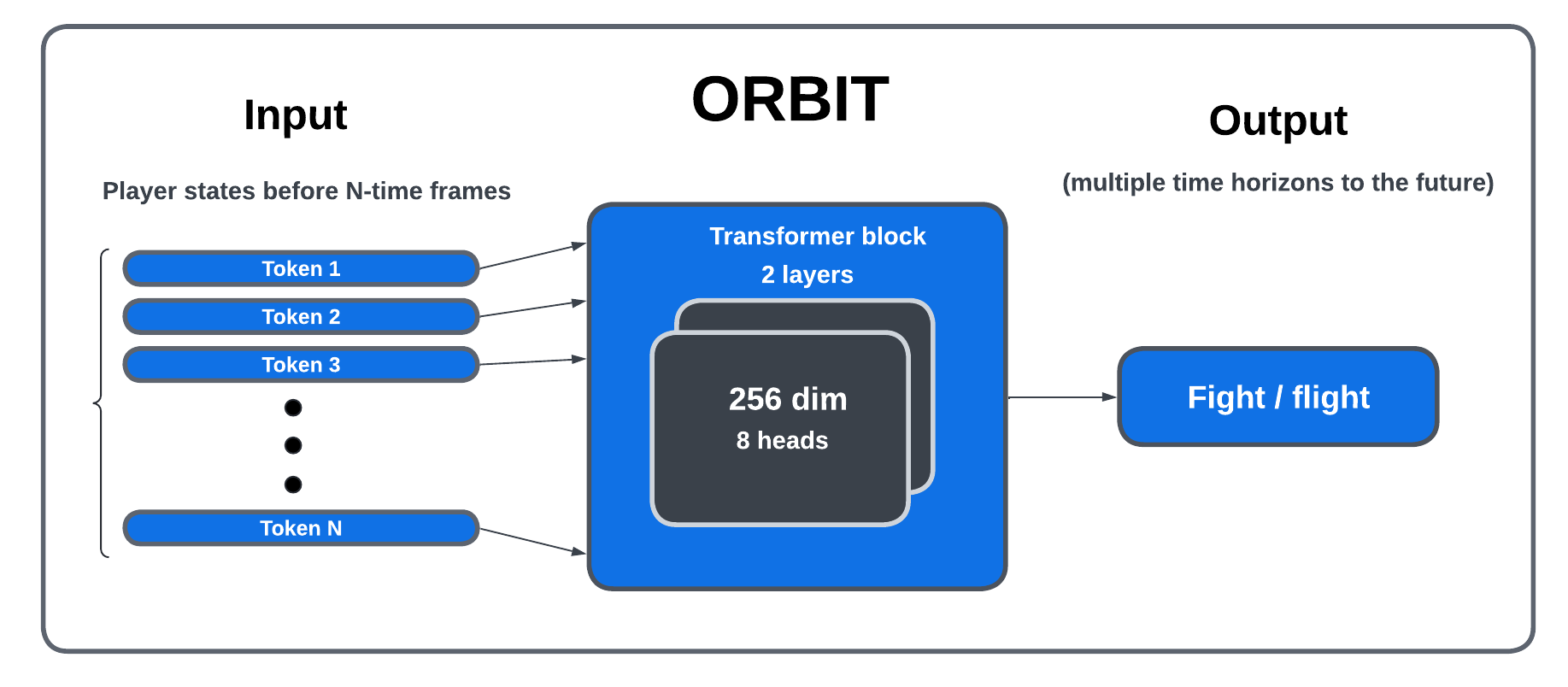}
    \caption{ORBIT architecture. A sequence of $N$ tick-level tokens, optionally preceded by global context tokens (map, player, rank), is processed by a Transformer encoder; the final representation feeds a classification head that predicts the behavioral response.}
    \label{fig:orbit_model_architecture}
\end{figure}

Global context tokens carry sequence-level information (map, analyzed player, and rank: learned embeddings for map and player, a projection of the normalized rank) and are prepended to the tick-level tokens. Sinusoidal time embeddings retain temporal order, and a Transformer encoder applies self-attention across all tokens. The encoder output at the final tick represents the classified situation and is passed to a classification head that produces the \textit{fight-or-flight} prediction (Fig.~\ref{fig:orbit_model_architecture}). Hyperparameters are given in the experimental setup.

The same architecture serves both experimental settings; only the temporal extent of the input differs.

\subsection{Gradient-Boosted Trees (EMBER)}
\label{subsec:EMBER}
EMBER (\emph{Ensemble Model for Behavioral Event Response}) is a gradient-boosted decision-tree model, the second proposed architecture. Unlike ORBIT, it operates on a temporally subsampled representation of the reconstructed Ego-Knowledge feature space, testing how much of the signal structured game-state features capture without the cost and complexity of a transformer. Gradient boosting was preferred over linear or logistic regression because of the high dimensionality and heterogeneous structure of the feature space; we use XGBoost~\cite{chen2016xgboost}.

Because the telemetry changes slowly between consecutive ticks, the native 64 samples per second were reduced to a coarser resolution, typically two per second. Each training instance consisted of a sequence of player-centric game-state snapshots spanning a fixed temporal window relative to the anchor tick (a sub-interval of $[-5\,\mathrm{s}, +1\,\mathrm{s}]$) at this reduced time step (typically and most commonly $0.5\,\mathrm{s}$ time step on interval $[-2\,\mathrm{s}, +0\,\mathrm{s}]$). The model is a binary classifier over \emph{fight} and \emph{flight}.

Hyperparameters (Table~\ref{tab:hyperparameters}) were kept close to library defaults as a simple, reproducible first configuration; tuning was explored but yielded a validation-accuracy range of only approximately $0.3$ percentage points, considered negligible at this stage.

\begin{table}[ht]
\centering
\begin{tabular}{lr}
\hline
\textbf{Parameter} & \textbf{Value} \\
\hline
\texttt{LEARNING\_RATE} & 0.02 \\
\texttt{MAX\_DEPTH} & 8 \\
\texttt{MIN\_CHILD\_WEIGHT} & 5.0 \\
\texttt{SUBSAMPLE} & 0.8 \\
\texttt{COLSAMPLE\_BYTREE} & 0.8 \\
\texttt{REG\_LAMBDA} & 1.0 \\
\texttt{REG\_ALPHA} & 0.0 \\
\texttt{MAX\_BIN} & 128 \\
\texttt{NUM\_BOOST\_ROUND} & 50000 \\
\texttt{EARLY\_STOPPING\_ROUNDS} & 1000 \\
\hline
\end{tabular}
\caption{Hyperparameter configuration of EMBER}
\label{tab:hyperparameters}
\end{table}

\subsection{Baseline Models}
\label{sec:baselines}

ORBIT and EMBER (Sections~\ref{sec:transformer} and \ref{subsec:EMBER}) are compared against two baselines of increasing capability: a majority-class classifier, a trivial lower bound relying solely on the marginal label distribution, and general-purpose large language models, which probe whether off-the-shelf models can solve the task from textual situation descriptions without task-specific architecture or fine-tuning. The baselines share the label definitions, input information, and metrics of ORBIT/EMBER (Section~\ref{sec:metrics}). The LLM baselines are scored on a fixed sample of 1{,}000 validation rows; for like-for-like comparison, EMBER and ORBIT are additionally scored on the same 1{,}000 rows (Section~\ref{sec:comparison}). \textit{Jev}, a decision model that does not generate text, is scored on the same 1{,}000 rows as an additional reference point (Section~\ref{sec:llm-setup}).

\subsubsection{Majority-Class Classifier}

The majority-class classifier uses no input features and always predicts the most frequent training class, verifying that the learned models improve over the prior class distribution. In the current split the majority class represents 51.5\% of the labeled situations, so this baseline achieves an expected accuracy of 51.5\% on data with the same class distribution. Since the classes are close to balanced, it is only slightly above random guessing.

\subsubsection{Language Model Baselines}
LLMs provide a strong and reproducible baseline for the same binary \emph{Fight} versus \emph{Flight} task without task-specific architecture or fine-tuning, and give practical reference points for performance, cost, and deployment complexity. The situation (features in Table~\ref{tab:dataset_features_all}) is encoded in three text-based representations:

\begin{itemize}
    \item \textbf{Narrative}: each game-state row is converted into concise natural-language situation lines by deterministic rule-based templates, using only the top-$k$ features of a feature-importance ranking computed on the training split (Fig.~\ref{fig:orbit-global-feature-importance-fight-all-features}); numeric values become interpretable phrases, missing values (including $-1$) are treated as absent, and few-shot prompts hold class-balanced labeled examples followed by the unlabeled query.
    \item \textbf{Compact JSON}: the same top-$k$ features are serialized as a minified, nested JSON object per example (e.g. \texttt{enemy\_X}); few-shot examples include the target \texttt{label}, the query omits it, and examples are newline-separated for token efficiency.
    \item \textbf{Full CSV}: all feature columns are serialized as CSV after removing metadata fields (e.g. \texttt{demo\_id}, \texttt{tick}) to prevent information leak; training examples carry a leading \texttt{label} column, then a header row and one data row per example. This is the textual representation closest to ORBIT's input.
\end{itemize}

\subsection{Evaluation Protocol}
\label{sec:evaluation}
Accuracy is the main comparative metric, since the dataset is well balanced and neither error type is preferred (metrics are defined in Section~\ref{sec:metrics}). Because querying the providers over the full validation set is prohibitively expensive, the LLM baselines are scored on a fixed sample of $1{,}000$ validation rows, which includes all $18$ gold-standard consensus moments that fall in the validation partition. ORBIT and EMBER are evaluated on the full validation set and, for like-for-like comparison, on the same $1{,}000$ rows; differences between paired predictions on those rows are tested with McNemar's test. No confidence intervals are reported for the headline accuracies (Section~\ref{sec:limitations}).

We also tested whether the models learn different players' characteristics. In the \emph{player-disjoint test}, only the 50 core players were kept; 10 formed the test set and the remaining 40 players' moments were used for training and validation. In the \emph{extended player-disjoint test}, the test set again comprised 10 core players while training and validation contained all other players (40 core players plus non-core players).

\section{Experimental Setup}
\label{sec:exp-setup}

This section specifies the data splits, hardware, metrics and model configurations behind the results in Section~\ref{sec:discussion}.

\subsection{Data Splits}
In the player-disjoint setup the test set contained 8{,}730 observations, with 16{,}609 and 2{,}993 rows in the train and validation sets respectively. The extended player-disjoint experiment used the same test set, with train and validation sets of 223{,}169 and 39{,}642 rows. Because early player-disjoint experiments found validation and test accuracies to be comparable (Section~\ref{sec:discussion}), the remaining experiments used a train/validation split only, without a dedicated test set, comprising 230{,}547 and 40{,}994 observations respectively. Table~\ref{tab:dataset_splits} summarizes the set sizes.

\begin{table}[h]
\centering
\small
\setlength{\tabcolsep}{3pt}
\begin{tabular}{lrrr}
\hline
\textbf{Setup} & \textbf{Train} & \textbf{Val.} & \textbf{Test} \\
\hline
Player-disjoint & 16{,}609 & 2{,}993 & 8{,}730 \\
Extended player-disjoint & 223{,}169 & 39{,}642 & 8{,}730 \\
Train/validation only & 230{,}547 & 40{,}994 & -- \\
\hline
\end{tabular}
\caption{Number of rows in train, validation, and test splits.}
\label{tab:dataset_splits}
\end{table}

\subsection{Hardware and Software}
All experiments were carried out on a single workstation with an NVIDIA RTX PRO 6000 Blackwell GPU (96~GB VRAM) and an Intel Xeon Gold 5412U CPU (24 cores, 48 threads), running Ubuntu 24.04 LTS with Python~3.13. ORBIT was implemented in PyTorch~2.11 and EMBER used XGBoost~3.2~\cite{chen2016xgboost}. Data preparation and analysis relied on Polars~1.31 and PyArrow~22.0 for columnar processing, the Delta Lake Python bindings (\texttt{deltalake}~1.3) for the medallion-architecture tables of Section~\ref{sec:dataset}, and NumPy, pandas, SciPy and scikit-learn for numerical computation and evaluation metrics; experiment tracking used Weights~\&~Biases. The language-model baselines were not run on this workstation: commercial models were accessed through their provider APIs and the open-source LLMs via cloud-hosted inference on OpenRouter (Section~\ref{sec:llm-setup}).

\subsection{Evaluation Metrics}
\label{sec:metrics}
The models are evaluated with standard binary classification metrics, comparing the predicted class of each selected situation (\textit{fight} or \textit{flight}) with its ground-truth label. Because the dataset is approximately balanced between the two classes, accuracy is the primary evaluation metric; precision and recall are reported as complementary metrics, with \textit{fight} as the positive class.

For the LLMs we additionally recorded inference time and cost. Inference time, in milliseconds, was defined as the wall-clock duration from sending the request to receiving the final token of the response; cost was derived from response metadata and pricing information in provider documentation. These two metrics are not measured under like-for-like conditions and should not be read as a controlled latency benchmark. For the LLM strategies the recorded time is an end-to-end wall-clock duration that includes the full client-to-provider round trip (network latency, provider-side queueing and server-side batching), none of which reflects the model's raw forward-pass cost, whereas ORBIT is timed as a single-sample local inference on the dedicated hardware described above, without any network component. The resulting difference is therefore one of \emph{orders of magnitude} and is dominated by deployment topology rather than by model size or computational complexity alone. We report these numbers because they characterize the operational reality of a locally served telemetry-native model versus an API-served general-purpose LLM, not as a compute-matched comparison; a strict compute-matched comparison, running the LLM baselines locally under identical batching and hardware conditions where possible, is left to future work.

\subsection{Model Configurations}

\subsubsection{ORBIT}
ORBIT, the main sequence-based neural architecture, is described in Section~\ref{sec:transformer}; only the concrete settings are reported here. The model was trained as a binary classifier for the \textit{fight-or-flight} task using cross-entropy loss, with the full set of hyperparameters summarized in Table~\ref{tab:orbit_hyperparameters}. Since ORBIT operates on temporal windows of tick-level tokens, the sequence history length was treated as an experimental parameter: windows of 7, 14, 21, 32, 64 and 192 ticks were evaluated, each ending at the anchor tick of the classified situation.

\begin{table}[t]
\centering
\caption{ORBIT hyperparameters used in the experiments.}
\label{tab:orbit_hyperparameters}
\begin{tabular}{l l}
\toprule
\textbf{Parameter} & \textbf{Value} \\
\midrule
Task & Binary classification \\
Target classes & \textit{fight}, \textit{flight} \\
Loss function & Cross-entropy loss \\
Token dimension & 256 \\
Transformer layers & 2 \\
Attention heads & 8 \\
Dropout & 0.3 \\
Sequence-history ticks & 7, 14, 21, 32, 64, 192 \\
Optimizer & AdamW \\
Learning rate & $1 \times 10^{-4}$ \\
Weight decay & $1 \times 10^{-5}$ \\
Maximum gradient norm & 1.0 \\
Training batch size & 512 \\
Validation batch size & 512 \\
Number of epochs & 50 \\
\bottomrule
\end{tabular}
\end{table}

\subsubsection{EMBER}
The EMBER hyperparameters are listed in Table~\ref{tab:hyperparameters}; any parameter not mentioned there was left at its default. Its main experimental parameter was the time step between subsequent feature values, since, unlike a transformer, XGBoost cannot inherently model temporal sequences and treats the input as a fixed set of feature columns.

\subsection{Language Model Setup}
\label{sec:llm-setup}
The LLM baseline suite evaluates multiple inference providers, comparing cloud-hosted, aggregated and local deployment paths under a consistent prompting and scoring protocol. Models employ prompt caching to substantially lower overall costs and use $temperature = 0$ to reduce additional hidden reasoning in the answer. Sample prompts for each game-state encoding are provided in Appendix~\ref{sec:appendix-llm-prompt-examples}, and the exact serialization of data into LLM input is described in Figure~\ref{fig:llm-serialization-pipeline}. The following providers were used:

\begin{itemize}
    \item \textbf{OpenAI cloud API:} GPT-5.6 (Terra), accessed through the OpenAI SDK with a medium reasoning effort.
    \item \textbf{OpenRouter cloud:} a cloud aggregation layer directing requests to upstream open-source model providers (via the OpenAI SDK); the evaluated models are Qwen~3.5~9B and Qwen~3.8~27B, with extra reasoning tokens turned off and context budgeted locally using \texttt{tiktoken}. The complete set of evaluated configurations is given in Table~\ref{tab:cross_model_comparison_table}.
    \item \textbf{Google Gemini cloud API:} Gemini~3~Flash and Gemini~3.7~Flash, accessed through the GenAI client with medium thinking effort.
    \item \textbf{Ollama local inference:} the lightweight local models \texttt{qwen2.5:0.5b} and \texttt{llama3.2:1b}, served through the local endpoint on port \texttt{11434} and run sequentially to avoid overwhelming the available GPU; later open-source evaluations moved to OpenRouter, allowing larger and more capable models to be tested through cloud-hosted routes.
\end{itemize}

The number of few-shot examples in the system prompt was bounded for cost. For the commercial models most runs used one eighth of the maximum example count that fits the model's context window (denoted 1/8 in Table~\ref{tab:cross_model_comparison_table}), with single runs at one quarter and three quarters of that maximum as checks; the open-source models used a fixed 32 examples.

As an additional reference point we evaluated \textit{Jev} (TypeSafe~AI), released in September 2026.\footnote{\url{https://typesafe.ai}; documentation at \url{https://docs.typesafe.ai}.} Its developer describes it as a ``System One'' model: instead of generating text token by token, it returns typed, probabilistic decisions in a single pass, and it is aimed at real-time loops such as games, robotics, and simulation, which makes it a relevant comparison for a telemetry task. Jev exposes three decision primitives: \textit{Choice} (a selection among predefined options), \textit{Score} (a position on an ordinal scale), and \textit{Noul} (a single yes/no probability). For this binary task, Choice selects between the two labels, Score is used with a two-level scale so that it returns the probability of \textit{fight}, and Noul answers whether the player will fight. We searched over the situation encodings, the number of labeled examples (from none to 120, bounded by the model's context limit), and the three primitives, scoring every configuration on the same 1{,}000 validation rows as the LLM baselines; Table~\ref{tab:cross_model_comparison_table} reports the best configuration.

\section{Results}
\label{sec:discussion}
This section reports the empirical findings, beginning with a descriptive validation of the extracted reaction-time structure (Section~\ref{sec:rt-results}) and then turning to the predictive results, their cross-model comparison, an explainability analysis, and qualitative case studies. The predictive results are organized around two temporal regimes that share the same ORBIT architecture but differ in what information the model is given relative to the reaction moment. In the \emph{Pre-Reaction Prediction} setting (Section~\ref{sec:per-model}), the model observes only information available up to and including the anchor tick; no part of the response is visible to it, so the task is genuine prediction of the eventual \textit{fight-or-flight} outcome. This is the stricter regime and serves as a lower bound on what pre-stimulus context alone can reveal.

In the \emph{Early Response Recognition} setting (Section~\ref{sec:early-response-results}), the input is extended by a short post-anchor window, so the model observes the first moments of the reaction as it begins to unfold. We regard this regime as the operationally decisive one. All analysis is performed \emph{ex post} on recorded demo files, and the objective is to recognize, as early as possible, which response is already in progress from its earliest kinematic signature: the model is detecting and classifying an event whose onset is already present in the signal rather than estimating a hidden future. The two regimes should therefore be read as complementary stages along the reaction timeline rather than as competing accuracy figures on a single problem.

\subsection{Descriptive Validation: Reaction-Time Structure of Back-Damage Events}
\label{sec:rt-results}
As a first application of the extraction procedure (Section~\ref{sec:reaction-extraction}), and as an initial validation of the dataset, we apply the reaction-detection protocol to back-damage events. The aim is deliberately modest: rather than fitting a multivariate model, we ask whether the extracted latencies form a structured, non-trivial distribution. Back-damage events constitute a natural surprise paradigm in which the threat arrives abruptly, the player is typically not visually prepared, and the canonical response is a rapid orientation correction. If such events yield a stable and interpretable latency distribution after filtering, the telemetry demonstrates one of the project's central requirements: that it preserves short-timescale behavioral organization rather than only coarse tactical outcomes.

\begin{figure}[h]
    \centering
    \includegraphics[width=0.9\linewidth]{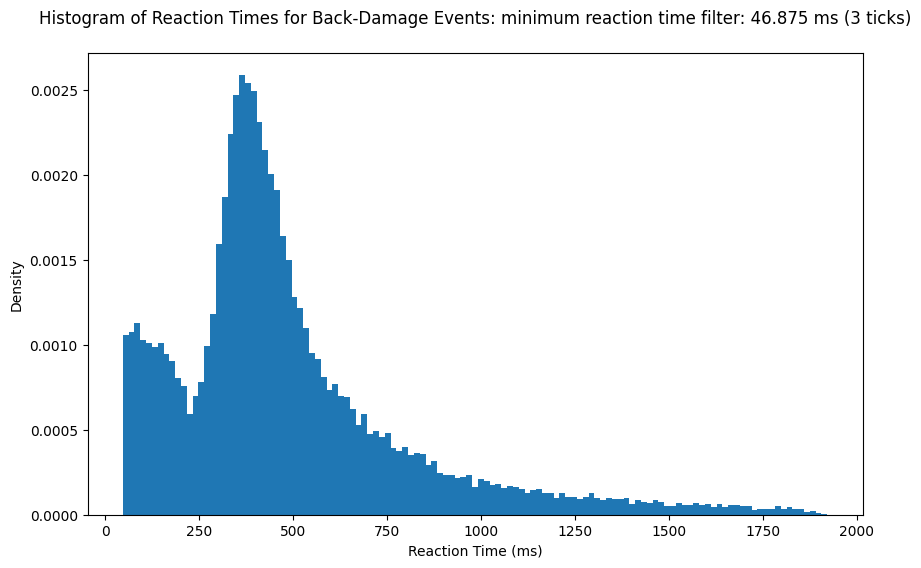}
    \caption{Histogram of reaction times for events in which the player is shot in the back. Base data granularity is 15.625~ms. The distribution provides an initial descriptive view of short-latency behavioral structure; number of situations is 20{,}114.}
    \label{fig:backshot-hist}
\end{figure}

\begin{figure}[h]
    \centering
    \includegraphics[width=0.9\linewidth]{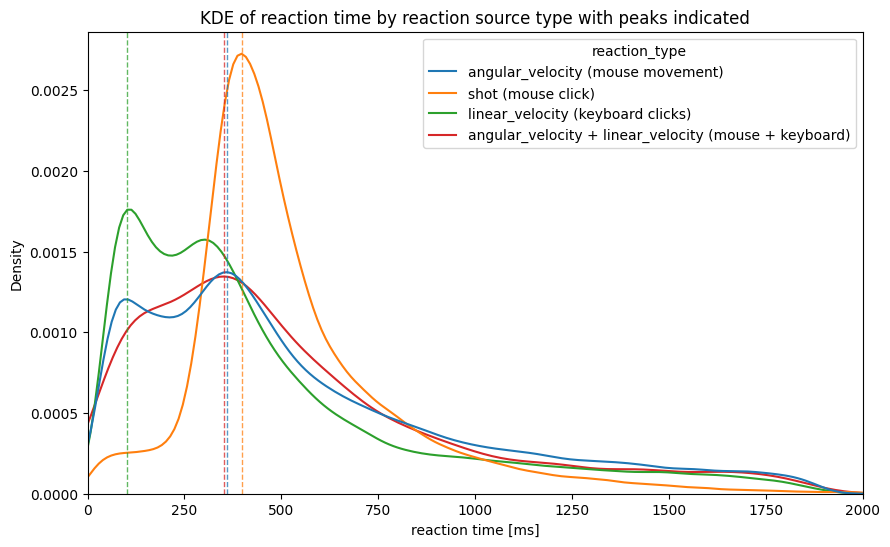}
    \caption{KDE plots comparing reaction times based on different sources of reaction (KDE plot follows the source histogram).}
    \label{fig:kdes}
\end{figure}

The resulting distribution (Fig.~\ref{fig:backshot-hist}) is neither uniform nor randomly dispersed. Three regions are visible: a pronounced concentration of responses near the lower measurable boundary, a broader dominant mass in the few-hundred-millisecond range, and a long, decreasing tail toward slower responses. The mid-latency mass plausibly corresponds to full defensive reorientation episodes, in which the player must detect the threat, update directional belief, initiate a turn, and prepare a follow-up action; the long tail plausibly captures delayed reactions, hesitation, or residual labeling ambiguity. The apparent separation into a faster and a slower band is the structure of primary interest, as it indicates that more than one timing regime is preserved in the data rather than collapsed into a single mode. Fig.~\ref{fig:kdes} compares the latency distributions obtained when the onset is detected from mouse movement (angular velocity), keyboard input (linear velocity), and weapon fire (shot). The three distributions are clearly offset: onsets detected from mouse movement and keyboard presses peak earliest, whereas the shot distribution is markedly right-shifted. This ordering matches the expected temporal structure of a motor reaction, in which a change in aim or movement is the first observable evidence of a response and firing is a later, downstream component. It reinforces our decision to anchor reaction onset on the first sudden kinematic change rather than on weapon fire (Section~\ref{sec:reaction-extraction}), since a fire-based anchor would systematically overestimate the true reaction latency.

The central observation is that the extraction pipeline preserves meaningful variability across fast, intermediate, and slow response bands rather than reducing them to a narrow artifact. This establishes, descriptively, that the source telemetry contains analyzable short-latency structure close to the project's operational fast-response regime. The interpretation of this distribution relative to the proposed 60--120~ms window, together with the claims it does and does not support, is taken up in Section~\ref{sec:discussion-conceptual}.

\subsection{Pre-Reaction Prediction Model Results}
\label{sec:per-model}

\subsubsection{ORBIT}
In this setting the model uses only information available at or before the anchor tick, so the task is to predict the player's eventual \textit{fight-or-flight} response before it is observed.

ORBIT was trained with several predefined sequence-history lengths to evaluate how the amount of pre-anchor temporal context affects classification performance. Each model used the same architecture and training setup, while the number of preceding ticks in the input sequence was varied over 7, 14, 21, 32, 64, and 192 ticks, each window ending at the anchor tick. The resulting validation accuracies are shown in Table~\ref{tab:orbit_history_window_results}.

\begin{table}[t]
\centering
\caption{Validation accuracy of ORBIT models trained with different sequence-history lengths in the pre-reaction prediction setting.}
\label{tab:orbit_history_window_results}
\begin{tabular}{c c}
\toprule
\textbf{Sequence-history ticks} & \textbf{Accuracy (\%)} \\
\midrule
7 & 71.20 \\
14 & 70.98 \\
21 & 70.96 \\
32 & 70.58 \\
64 & 68.93 \\
192 & 68.32 \\
\bottomrule
\end{tabular}
\end{table}

The best validation accuracy, 71.20\%, was achieved with the shortest evaluated window of 7 ticks. Performance gradually decreased as the window increased, with the 192-tick model achieving 68.32\%. Although this trend suggests that shorter pre-anchor temporal context was more effective in the current experimental setting, the observed differences should not be interpreted as statistically significant without repeated runs or confidence-interval analysis. The validation set was used for this comparison because the history window is a hyperparameter.

\begin{figure}[t]
    \centering
    \includegraphics[width=1.0\linewidth]{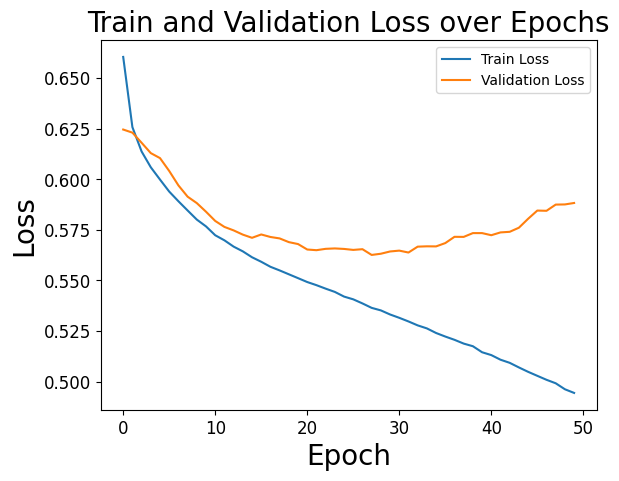}
    \caption{Training and validation loss of the selected 7-tick ORBIT model in the pre-reaction prediction setting over 50 epochs. The validation loss reaches its minimum around epoch 30 and then increases while the training loss continues to decrease, indicating overfitting in the later stages of training.\protect\footnotemark}
    \label{fig:orbit_7tick_loss}
\end{figure}

The remaining analysis focuses on the 7-tick ORBIT model. Fig.~\ref{fig:orbit_7tick_loss} shows its training and validation loss over 50 training epochs. The training loss decreases consistently throughout; the validation loss follows the same trend during the initial epochs, stops improving after approximately epoch 30, and then begins to increase while the training loss continues to fall, indicating overfitting in the later stages of training. The final evaluation therefore used the checkpoint with the lowest validation loss rather than the weights from the last epoch, so that the reported performance corresponds to the model state with the best observed validation generalization.

\footnotetext{We hypothesize that the early onset of overfitting is partly due to the limited size of the training set, since Transformer architectures are known to be data-hungry and to generalize substantially better as the training data grows \cite{dosovitskiy2021image,kaplan2020scaling}.}

The selected 7-tick ORBIT model achieved a validation accuracy of 71.20\%, outperforming the majority-class baseline and all longer-history ORBIT variants. This suggests that, under the current dataset size and training setup, a short pre-anchor temporal context was sufficient for the model to capture useful information for \textit{fight-or-flight} prediction before the player's response was observed. To assess whether the model generalizes beyond players observed during training, the 7-tick model was also evaluated in a player-disjoint setting, in which the test set contained players not present in the training split. It achieved an accuracy of 71.1\%, very close to the 71.20\% obtained in the standard validation setting, suggesting that the model does not substantially lose predictive performance on unseen players and learns patterns of \textit{fight-or-flight} behavior that generalize beyond player-specific identity.

\subsubsection{EMBER}
Initial experiments confirmed that the constructed dataset and the labeling pipeline produce a consistent and learnable signal. A classifier was trained on temporally aggregated Ego-Knowledge features, with input samples sampled at a purposefully higher granularity of history (0.5~s) and incorporating a short history window of up to two seconds prior to the reaction moment. Despite the simplicity of the model and the absence of hyperparameter tuning, training was stable and converged reliably across runs. The best achieved validation accuracy was 73.5\%.

The player-disjoint test and its extended version indicate that the model is not learning a player-specific reaction prior and instead generalizes across players; this property can later be used to compare a player's decision process against a model-derived baseline. The player-disjoint test yielded a validation accuracy (data from the same players as the training set) of 70.7\%, while the test accuracy (players never seen in either training or validation) was 71.4\%. The extended player-disjoint test, conducted to check that adding more random players to the training and validation sets does not diminish this property, yielded validation and test accuracies of 73.3\% and 73.7\%, respectively. The slight increase of test over validation accuracy may reflect lower variability of source players in the test set (more players can introduce more outliers).

\subsubsection{Language Models}
The LLM baselines reach at most $62.7\%$ accuracy over 1{,}000 validation predictions (Table~\ref{tab:cross_model_comparison_table}); the configurations fall between $57.3\%$ and $62.7\%$, and the same model can differ by several points between encodings and example budgets. The best-performing configuration is \textit{Gemini 3.7 Flash} with medium thinking and Narrative encoding at $62.7\%$, followed by \textit{GPT-5.6 Terra} with Narrative encoding at $61.8\%$ and \textit{Gemini 3.7 Flash} with Full-Dump encoding and three quarters of the maximum example budget at $61.6\%$; the latter is the most expensive configuration in the comparison, so a larger example budget did not translate into higher accuracy. The open-source \textit{Qwen 3.5 9B} and \textit{Qwen 3.8 27B} models reach $59.0$--$59.9\%$ at a small fraction of the cost. An earlier evaluation on a previous revision of the dataset, using the model generation available at the time, produced a best LLM accuracy of $63.5\%$; the LLM ceiling has therefore not moved with a newer generation of models, which is consistent with the limitation lying in the modality rather than in model capability.

In the earlier evaluation, \textit{Centaur 70B}~\cite{binz2024centaurfoundationmodelhuman} and \textit{Minitaur 8B} were included as an additional comparison layer because they are explicitly designed to model human choice behavior. Centaur was trained on \textit{Psych-101}, a large collection of trial-level human responses from psychological experiments, and therefore represents a relevant baseline for evaluating whether a cognition-oriented foundation model transfers to the fight-versus-flight setting. Both models performed poorly on our game-state dataset, reaching only $48.3\%$ and $50.4\%$ accuracy on the previous dataset revision, respectively, and their predictions showed a strong bias toward the \textit{Flight} class; they were not re-evaluated on the current revision and are therefore not listed in Table~\ref{tab:cross_model_comparison_table}. A likely explanation is the mismatch between training domain and target representation: \textit{Psych-101} consists primarily of natural-language descriptions of controlled psychological experiments and participant choices, whereas our dataset contains high-dimensional, temporally structured game-state observations with 154 numerical and categorical features describing health, movement, visibility, enemy proximity, aim alignment, and recent combat dynamics. These results suggest that cognition-oriented pretraining alone is insufficient for this setting unless the model is adapted to the specific structure, feature semantics, and class balance of the game-behavior dataset.

We additionally examined whether a very small open-source model could be adapted to the task via parameter-efficient fine-tuning: \textit{Qwen2.5-0.5B} was fine-tuned on our labeled game states using LoRA (rank $8$, learning rate $2\times10^{-4}$), framing the problem as a JSON-to-JSON mapping with no class weighting. The result was clearly negative: although the model learned to reproduce the expected JSON output structure, it collapsed to a constant prediction, emitting the \textit{Flight} class for every input, so precision, recall, and $F_1$ score were all effectively $0$. We conclude that sub-billion-parameter models of this scale lack the capacity to recover a meaningful decision boundary for this setting.

The best \textit{Jev} configuration (Score primitive, Narrative encoding restricted to 20 features, 104 labeled examples) reaches $63.4\%$ accuracy on the same 1{,}000 rows, with precision $61.91\%$ and recall $68.88\%$, at $0.48$\,s and \$1.27 per 1{,}000 predictions; on the same input the Noul and Choice primitives reach $62.3\%$ and $61.9\%$. Across the configurations tried, accuracy ranges from $47.3\%$ with no examples to $63.4\%$. Jev therefore sits at the top of the baselines that read a textual description of the situation, $0.7$ percentage points above \textit{Gemini 3.7 Flash} at roughly a quarter of its cost, and well below the models that consume the telemetry directly. It is not part of the paired tests in Section~\ref{sec:comparison}; on the same rows EMBER and ORBIT lead it by $10.2$ and $7.5$ percentage points.

\subsubsection{Majority-Class Baseline}
The majority-class baseline yields accuracy equal to the percentual representation of the most common class, 51.5\%, and gives perspective on how much the machine learning models improve on this naive baseline.

\subsection{Pre-Reaction Prediction Cross-Model Comparison}
\label{sec:comparison}

EMBER reaches $73.5\%$ accuracy and ORBIT $71.2\%$ on the full validation set; the strongest LLM configuration, \textit{Gemini 3.7 Flash} with medium thinking and Narrative encoding, reaches $62.7\%$ over 1{,}000 predictions drawn from the same validation set (Table~\ref{tab:cross_model_comparison_table}). TabPFN provides a substantially stronger non-generative baseline: \textit{TabPFN v3.5} reaches \(71.5\%\) accuracy with a \(25{,}000\)-sample context, \(72.6\%\) with \(100{,}000\), and \(73.5\%\) with the full \(200{,}000+\) context, matching EMBER on overall accuracy, but providing less balance between Type I and Type II errors. To compare like for like, EMBER and ORBIT were scored on the same 1{,}000 rows, where they reach $73.6\%$ and $70.9\%$, respectively. McNemar's test on the paired predictions gives $p = 1.3\times10^{-9}$ for EMBER and $p = 1.5\times10^{-6}$ for ORBIT against the strongest LLM configuration, and $p = 1.1\times10^{-23}$ and $p = 1.3\times10^{-17}$ against the majority-class baseline. The margin over the best LLM, roughly eleven percentage points for EMBER and eight for ORBIT, is therefore significant on identical trials under a test that treats the rows as independent (Section~\ref{sec:limitations}).
Additionally, in its full context, \textit{TabPFN v3.5} reaches the same 73.5\% validation accuracy as EMBER, so the present results do not establish an accuracy advantage for EMBER over strong pretrained tabular models. Two caveats remain: the commercial LLMs were run with a bounded few-shot example budget (Section~\ref{sec:llm-setup}), and no confidence intervals are reported (Section~\ref{sec:limitations}).

Both ORBIT and EMBER also provide substantially better operational efficiency: for ORBIT, inference is approximately $1{,}412\times$ faster and $10{,}322\times$ cheaper than \textit{Qwen 3.5 9B} with Narrative encoding, the fastest and cheapest LLM configuration, corresponding to a $99.99\%$ cost reduction; against the most accurate LLM configuration ($1.81$\,s and \$4.70 per 1{,}000 predictions versus $0.34$\,ms and \$0.000062 for ORBIT) the ratios are roughly $5{,}300\times$ and $76{,}000\times$. TabPFN occupies an intermediate operating point: the full-context \textit{TabPFN v3.5} reaches $73.5$\% accuracy at \(0.215\,\mathrm{s}\) and $0.0429$ per $1{,}000$ predictions, making it roughly $8.4\times$ faster and $110\times$ cheaper than the most accurate LLM configuration, while ORBIT remains approximately $630\times$ faster and $690\times$ cheaper than this TabPFN setting. EMBER achieves the greatest decrease in both cost and latency. \textit{Jev}, which targets low-latency decision loops, is among the faster and cheaper API-served options ($0.48$\,s and \$1.27 per 1{,}000 predictions), yet ORBIT remains roughly $1{,}400\times$ faster and $20{,}000\times$ cheaper. As noted in Section~\ref{sec:metrics}, this is not a compute-matched comparison: the API-served figures are end-to-end wall-clock measurements including network and queueing time, and the gap is dominated by deployment topology. Results from the initial experiments using locally hosted models were excluded from the comparison due to unstable inference, given the available GPU resources, with an extreme bias toward a single prediction class.

Because the models consume Ego-Knowledge data only up to the tick of the stimulus (visibility or damage event), and the dataset was constructed with a clear attempt to minimize other stimuli affecting the current player action, only player data (position, rotation, and so on) has a truly temporal character in these experiments. The part of the data that might be most influential on the player's decision (enemy position, for example) is essentially reduced to a single value. It is therefore not necessarily surprising that the boosted-tree architecture outperformed the transformer here. The learning process nevertheless worked well for the transformer as well, so for related future use cases with data of a more temporal character this architecture has a solid base and may later prove more valuable.

Table~\ref{tab:cross_model_comparison_table} presents the comparison of all models on the chosen factors. Overall, the results suggest that the predictive signal in this task is contained primarily in the structured telemetry rather than in its natural-language representation. While large language models are able to recover part of this signal, they are consistently outperformed by models designed to operate directly on tabular inputs. Notably, EMBER matches the accuracy of the strongest foundation-model configuration while requiring only a fraction of its inference cost and latency, indicating that substantially greater model complexity does not translate into a corresponding performance advantage for this setting.

\begin{table*}[t]
    \caption{Performance comparison of prediction strategies; the LLM and Jev rows are over 1,000 predictions from the current validation set.}
    \label{tab:cross_model_comparison_table}
    \centering
    \begin{tabular}{lrrrrr}
        \hline
        \textbf{Strategy} & \textbf{Accuracy (\%)} & \textbf{Avg. time (s)} & \textbf{Total Cost (\$)} & \textbf{Precision (\%)} & \textbf{Recall (\%)} \\
        \hline
        Gemini 3.7 Flash Medium Narrative (1/8)   & 62.7 & 1.81 & 4.70  & 59.66 & 77.51 \\
        Gemini 3.7 Flash Medium Full-Dump (3/4)   & 61.6 & 6.31 & 60.03 & 58.19 & 81.33 \\
        Gemini 3 Flash Medium Full-Dump (1/8)     & 59.7 & 2.97 & 14.72 & 56.81 & 79.52 \\
        Gemini 3 Flash Medium Full-Dump (1/4)     & 57.3 & 3.28 & 15.97 & 55.42 & 72.89 \\
        Gemini 3 Flash Medium Full-Dump (3/4)     & 57.4 & 2.10 & 40.11 & 55.36 & 74.70 \\
        Gemini 3 Flash Medium Narrative (3/4)     & 60.3 & 5.07 & 16.98 & 57.69 & 76.10 \\
        Gemini 3 Flash Medium Compact (3/4)       & 61.1 & 4.74 & 23.64 & 58.61 & 74.50 \\
        \hline
        GPT-5.6 Terra Medium Full-Dump (1/8)      & 57.7 & 3.32 & 29.76 & 55.71 & 73.49 \\
        GPT-5.6 Terra Medium Narrative (1/8)      & 61.8 & 1.42 & 14.21 & 59.48 & 73.09 \\
        \hline
        Qwen 3.8 27B Full-Dump (32)               & 59.9 & 1.13 & 4.85  & 59.34 & 61.85 \\
        Qwen 3.8 27B Narrative (32)               & 59.0 & 0.92 & 2.18  & 62.09 & 45.38 \\
        Qwen 3.5 9B Full-Dump (32)                & 59.4 & 0.82 & 1.72  & 56.32 & 82.33 \\
        Qwen 3.5 9B Narrative (32)                & 59.0 & 0.48 & 0.64  & 61.52 & 47.19 \\
        \hline
        Jev (Score) Narrative, 20 features (104) & 63.4 & 0.48 & 1.27 & 61.91 & 68.88 \\
        \hline
        TabPFN v2, context 25,000$^{\ddagger}$ & 67.7 & 0.028 & 0.0057 & 62.9   & 80.2 \\
        TabPFN v3.5, context 25,000$^{\ddagger}$ & 71.5 & 0.0089 & 0.0018 & 66.9   & 80.1 \\
        TabPFN v3.5, context 100,000$^{\ddagger}$ & 72.6 & 0.061 & 0.0122 & 68.7   & 78.2 \\
        TabPFN v3.5, full context (200,000+)$^{\ddagger}$ & \textbf{73.5} & 0.215 & 0.0429 & 69.6   & 79.0 \\
        \hline
        Majority-class baseline*                 & 51.5 & -    & -     & NaN   & 0 \\
        \textbf{EMBER}*                           & \textbf{73.5} & \textbf{0.00001} & \textbf{0.000002} & \textbf{70.97} & 76.12 \\
        ORBIT*                                    & 71.2 & 0.00034 & 0.000062 & 67.0 & \textbf{79.5} \\
        \hline
    \end{tabular}
    \vspace{2mm}
\begin{minipage}{0.8\linewidth}
\footnotesize
*EMBER, ORBIT, and majority-class accuracies are over the full validation set; on the 1,000 rows used for the LLM evaluation they are 73.6\%, 70.9\%, and 50.2\%, respectively (paired tests in the text). The 1,000 rows subset contains 502 flight and 498 fight instances. The value in parentheses gives the few-shot example budget: a fraction denotes the share of the maximum example count that fits the model's context window, and an integer denotes a fixed number of examples. $^{\ddagger}$ For TabPFN, context rows are chosen as a random subset from EMBER and ORBIT training set. Random validation subset contained 52.3\% of the majority (flight) class. For full comparability, full context run (all training rows) is evaluated also on all validation rows. 
\end{minipage}
\end{table*}

\subsection{Early Response Recognition Results}
\label{sec:early-response-results}

In the second experimental setting, we evaluated ORBIT and EMBER in the \emph{Early Response Recognition} task defined in Section~\ref{subsec:early-response-recognition}, in which the input sequence may include a short post-anchor window, so that the model observes an increasing portion of the player's actual reaction.

\subsubsection{ORBIT}

\begin{figure}[t]
    \centering
    \includegraphics[width=1.0\linewidth]{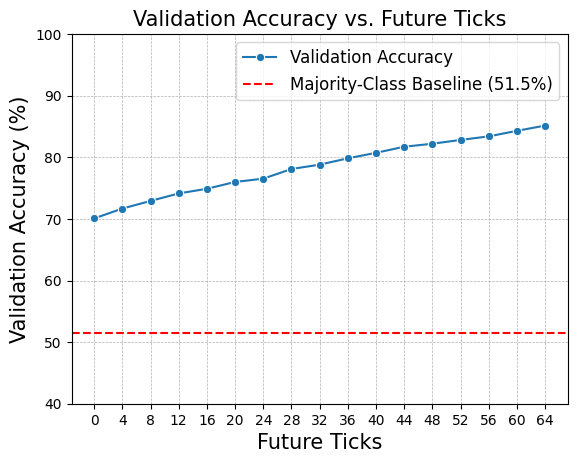}
    \caption{Validation accuracy of ORBIT in the early response recognition setting as the number of future ticks increases. The history window is fixed to 7 ticks, while the future window is extended up to 64 ticks, corresponding to approximately one second after the anchor moment.}
    \label{fig:early_response_future_window_accuracy}
\end{figure}

The history window was fixed to 7 ticks, the best configuration in the pre-reaction prediction experiment, and ORBIT models were trained with progressively longer future windows after the anchor tick: 4, 8, 12, and more future ticks, up to 64. Since CS2 demos are sampled at 64 ticks per second, the maximum future window corresponds to approximately one second after the anchor moment, matching the label definition. Accuracy increases consistently as more post-anchor ticks are included (Fig.~\ref{fig:early_response_future_window_accuracy}), reaching 85.15\% with the full one-second window.

The monotonic improvement is expected, because the label is itself defined by the behavioral response shortly after the anchor moment: as the model receives more of the unfolding reaction, the task gradually shifts from predicting the player's likely response before it is visible to recognizing it from partial evidence of the behavior itself. These results should therefore be interpreted differently from the pre-reaction results. The experiment does not evaluate whether the model can predict a response before it starts, but how quickly and reliably ORBIT identifies the response once its first part becomes observable, and the consistent increase suggests that even a short post-anchor context helps.

\subsubsection{EMBER}

\begin{figure}[t]
    \centering
    \includegraphics[width=1.0\linewidth]{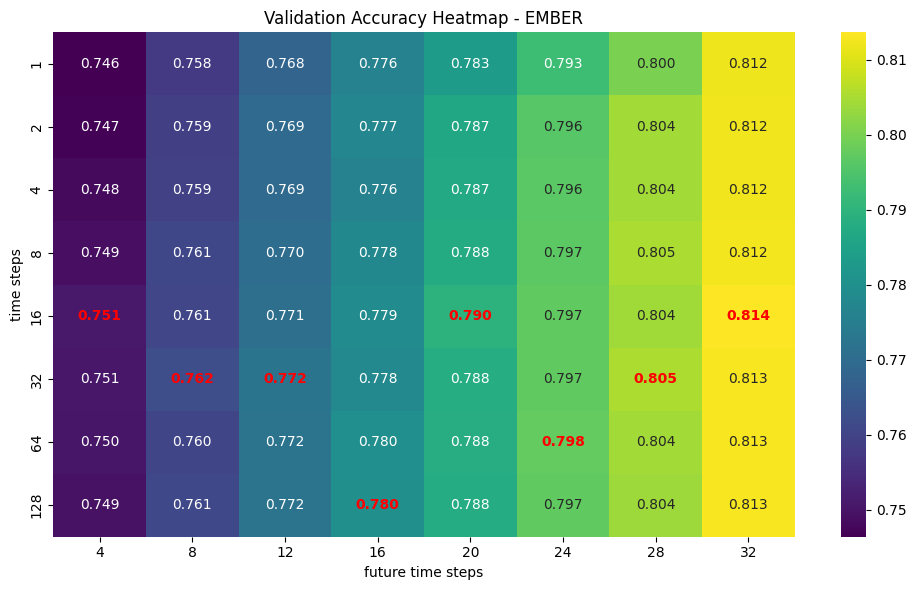}
    \caption{Validation accuracy of EMBER in the early response setting as a function of future time steps (maximum difference in ticks between stimulus and the latest moment in the model input) and time step (in ticks). The best accuracy for each future time step is highlighted in red.}
    \label{fig:ember_future}
\end{figure}

A similar experiment was carried out with EMBER, except that the whole history was used instead of the 7-tick history and training covered a grid of time steps and future ticks (the latter being the maximum tick difference from the stimulus, not necessarily the number of future ticks in the input). The results (Fig.~\ref{fig:ember_future}) indicate that EMBER performs better with a higher time step and improves with each forward extension of the time frame. Part of this gain can be attributed to the task shifting from purely predictive to partly classificational; however, the reaction-time histograms in Section~\ref{sec:rt-results} show that only a very small portion of reactions occurs in the first 4 or even 8 ticks. It is therefore more likely that this result points to the reaction not being fully correlated with intention, as later stimuli can change the final decision from the initial intention.

\subsection{Explainability}
\label{sec:xai}

\begin{figure*}[t]
    \centering
    \includegraphics[width=0.82\textwidth]{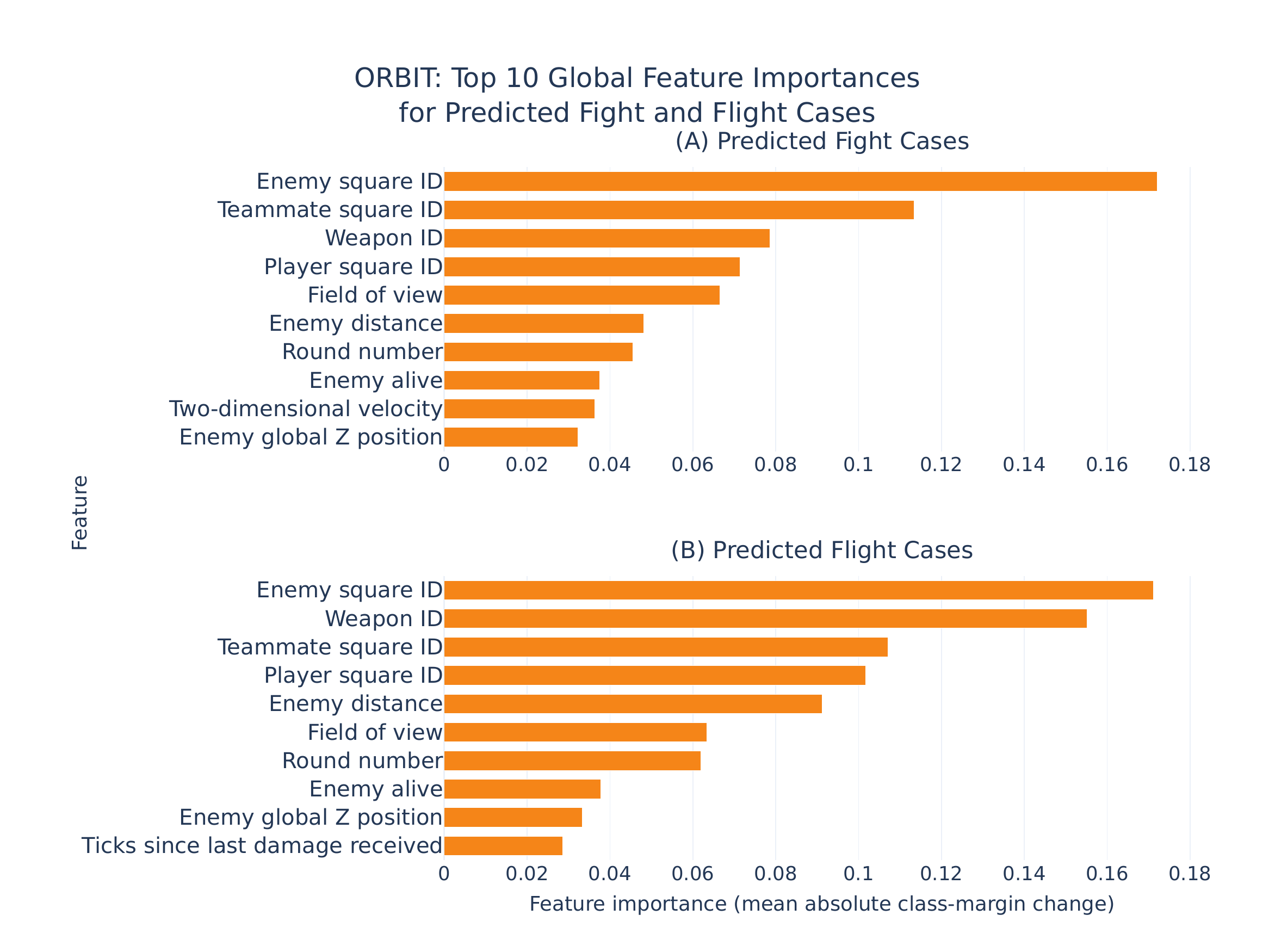}
    \caption{Global perturbation-based feature importance of ORBIT for samples predicted as \textit{fight} and \textit{flight}. Importance is measured as absolute change in predicted-class margin, so the plot should be interpreted as model sensitivity rather than causal evidence.}
    \label{fig:orbit-global-feature-importance-top10}
\end{figure*}

Feature importance was computed with a perturbation-based procedure: each feature was replaced with a neutral baseline value and the resulting change in the predicted-class margin was measured. Fig.~\ref{fig:orbit-global-feature-importance-top10} shows that, for both predicted \textit{fight} and predicted \textit{flight} cases, ORBIT is most sensitive to features describing the player state, spatial context, weapon state, field of view, enemy configuration, and recent combat information, suggesting that predictions are driven by the interaction between the player's local situation and the geometry of the nearby engagement rather than by a single isolated variable. \textit{Fight} predictions are more associated with the immediate engagement context, whereas \textit{flight} predictions also reflect cues related to threat and disengagement, such as recent damage history. Overall, the plots support the intended role of Ego-Knowledge as a compact representation of the local tactical context. Full rankings are given in Fig.~\ref{fig:orbit-global-feature-importance-fight-all-features} (ORBIT, predicted \textit{fight}) and Fig.~\ref{fig:xgb-feats-1} (EMBER) in the Appendix.

These results should still be interpreted as model sensitivity diagnostics rather than causal explanations. Some inputs, such as weapon and spatial identifiers, are represented through learned embeddings, so their importance reflects ORBIT's sensitivity to the learned representation rather than a direct physical intervention in the match.

In all LLM models, the main determinants of \textit{fight} situations are crosshair proximity, weapon readiness, and enemy awareness; six recurring factors appear in the models' reasoning (crosshair distance, weapon equipped, visibility and awareness, movement velocity, temporal cues such as recent firing and weapon switches, and health and team status), and Table~\ref{tab:fight-flight-patterns} in the Appendix lists the identified \textit{fight} and \textit{flight} patterns together with the error patterns behind prediction mistakes. Larger commercial models (OpenAI and Gemini, and DeepSeek in the earlier evaluation) prioritize tactical advantages and aim alignment, while smaller local models often collapse into repetitive templates or meta-commentaries; a recurring systemic bias is to predict \textit{fight} whenever an enemy is visible, often failing to recognize tactical retreats or ``saving'' behaviors.

\subsection{Label Validation by Majority Voting}
\label{sec:label_validation_majority_voting}

A subset of situations was selected for manual labeling as described in Section~\ref{sec:manual_labeling}, chosen so that external influences were expected to be limited and the timing of the stimulus was well defined. Each situation was labeled independently by seven annotators with advanced knowledge of CS2 gameplay, who could assign a response label or leave the situation unresolved when the available evidence was insufficient or ambiguous.

A gold-standard subset, containing exclusively damage events, was constructed by majority voting: a situation received a gold-standard label only if more than 50\% of annotators agreed on the same response class. Of the 200 manually inspected situations, 165 received a majority-vote label, a consensus coverage of 82.5\%. Consensus coverage is the proportion of moments on which a majority was reached; with seven annotators and a binary choice, a majority is near-certain unless annotators abstain, so it principally reflects abstention and is not an agreement statistic. The agreement statistic is Fleiss's kappa: 0.38 for the entire subset of 200 moments and 0.44 for the gold-standard set, corresponding to fair (0.38) and moderate (0.44) agreement on the conventional Landis and Koch interpretation. Agreement of this level bounds the accuracy achievable by any model trained against these labels, so the reported model accuracies should be read against that ceiling rather than against 100\%.

Table~\ref{tab:majority_vote_validation} summarizes the validation: agreement between the automatic labeling procedure and the gold-standard set is 80.6\%. A subset (119 instances) of situations with majority-vote label which also fit the dataset filtering conditions, used as input to EMBER, did not yield significantly different results from the rest of the validation set (74.8\%, which given the gold-standard set size is comparable to the whole-validation 73.5\%). The gold-standard set is subject to expansion: it comprises damage events only and does not yet include the majority event type, visibility, which poses more challenges for correct filtering.

\begin{table}[H]
\centering
\caption{Summary of majority-vote label validation. Agreement with the automatic labeling system is computed on the majority-vote gold-standard subset.}
\label{tab:majority_vote_validation}
\begin{tabular}{l c}
\toprule
\textbf{Quantity} & \textbf{Value} \\
\midrule
Manually inspected situations & 200 \\
Annotators per situation & 7 \\
Majority-vote threshold & $>50\%$ \\
Situations with majority-vote label & 165 \\
Number of situations used for EMBER validation & 119 \\
Consensus coverage & 82.5\% \\
Automatic label agreement with gold-standard set & 80.6\% \\
EMBER accuracy on gold-standard set & 74.8\% \\
\bottomrule
\end{tabular}
\end{table}

\subsection{Qualitative Case Studies}
\label{sec:case-studies}

To complement the quantitative evaluation, we analyze four qualitative case studies from the validation set, two correctly classified and two failures, all from the middle area of the map \textit{Mirage}. The analyzed player is the orange \textit{Terrorist} side player and the opponent the blue \textit{Counter-Terrorist} side player; for the two cases shown, the first image is the anchor moment at which the prediction was made and the second the same situation approximately one second later. These are illustrative examples, not a statistical evaluation.

\begin{figure*}[t]
    \centering

    \begin{subfigure}[t]{0.48\textwidth}
        \centering
        \includegraphics[width=\linewidth]{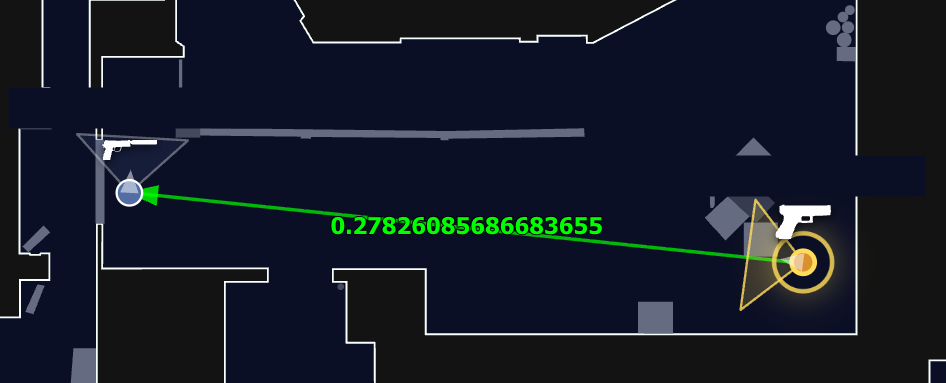}
        \caption{Anchor moment.}
        \label{fig:case_correct_fight_anchor}
    \end{subfigure}
    \hfill
    \begin{subfigure}[t]{0.48\textwidth}
        \centering
        \includegraphics[width=\linewidth]{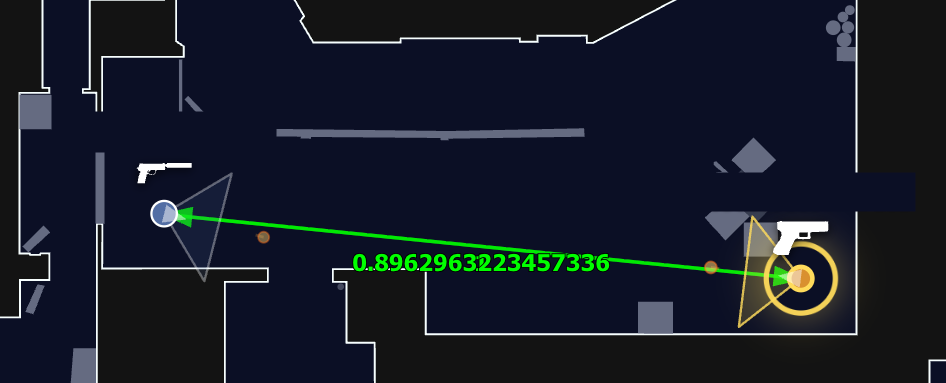}
        \caption{Approximately one second later.}
        \label{fig:case_correct_fight_followup}
    \end{subfigure}

    \caption{Correctly classified \textit{fight} example: the analyzed player spots the opponent on lower middle and commits to the duel.}
    \label{fig:case_correct_fight}
\end{figure*}

Fig.~\ref{fig:case_correct_fight} shows a correctly classified \textit{fight}. At the anchor moment (Fig.~\ref{fig:case_correct_fight_anchor}) the analyzed player spots the opponent on lower middle and, instead of disengaging, takes the duel (Fig.~\ref{fig:case_correct_fight_followup}). The example comes from the pistol round, where weapon strength is typically comparable, so committing to the \textit{fight} is a plausible tactical decision. In the correctly classified \textit{flight} case (not shown), the analyzed player, holding a lighter sniper rifle, identifies that the window is controlled by an opponent with an \textit{AWP} on a favorable angle and disengages, as the model predicts.

In the first failure case (not shown), the model predicted \textit{fight} for a player holding an \textit{AWP} against an \textit{M4 rifle}, a favorable long-range matchup, who after missing the first shot hides and reloads rather than re-engaging; the model appears to interpret the weapon advantage and initial engagement as evidence for \textit{fight} and fails to capture the subsequent disengagement.

\begin{figure*}[t]
    \centering

    \begin{subfigure}[t]{0.48\textwidth}
        \centering
        \includegraphics[width=\linewidth]{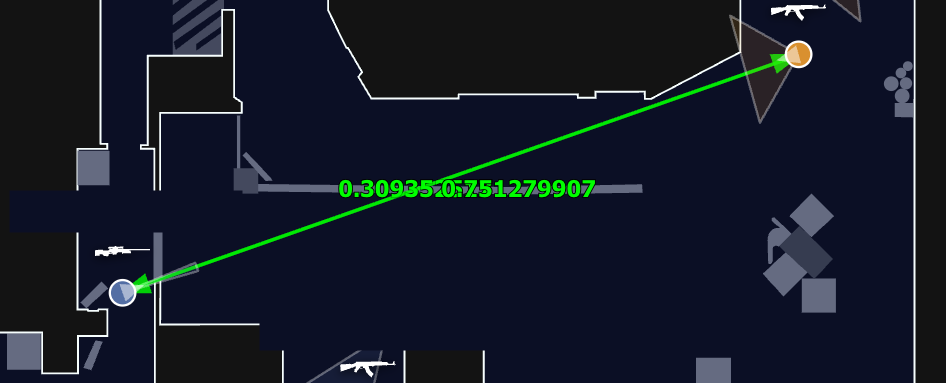}
        \caption{Anchor moment.}
        \label{fig:case_incorrect_flight_anchor}
    \end{subfigure}
    \hfill
    \begin{subfigure}[t]{0.48\textwidth}
        \centering
        \includegraphics[width=\linewidth]{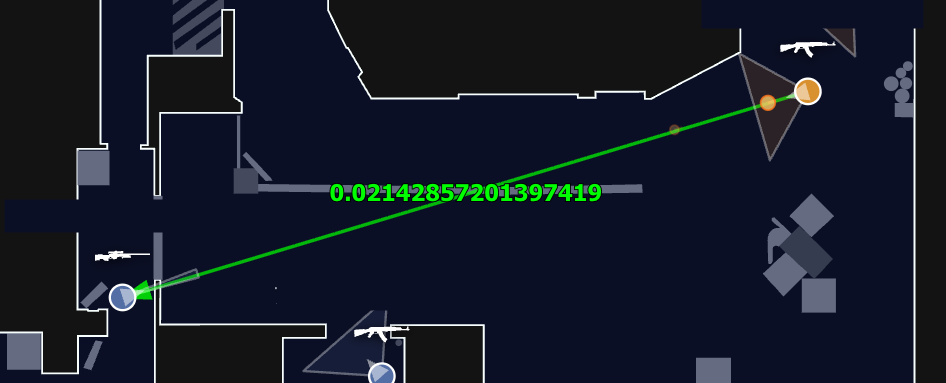}
        \caption{Approximately one second later.}
        \label{fig:case_incorrect_flight_followup}
    \end{subfigure}

    \caption{Incorrectly classified \textit{flight} example: the analyzed player, with an \textit{AK-47}, faces an \textit{AWP} in a window but continues the duel.}
    \label{fig:case_incorrect_flight}
\end{figure*}

Fig.~\ref{fig:case_incorrect_flight} shows the opposite failure: the model predicted \textit{flight}, but the player continued into the duel. At the anchor moment (Fig.~\ref{fig:case_incorrect_flight_anchor}) the analyzed player, with an \textit{AK-47}, is exposed to an \textit{AWP} in a window, a risky engagement because the \textit{AWP} player controls a strong angle and, even after missing the first shot, can often retreat behind cover, reload, and re-peek; the player nevertheless commits to the fight (Fig.~\ref{fig:case_incorrect_flight_followup}). Such cases are difficult because the correct label depends not only on the objective geometry and weapon matchup, but also on the player's willingness to take a high-risk duel.

Overall, the examples show that ORBIT can correctly identify both engagement and disengagement when the player's behavior is consistent with the local tactical context. The failure cases, however, suggest that some decisions depend on factors only partially captured by the current representation, such as risk preference, confidence in a weapon matchup, or the intention to re-peek after temporarily taking cover.

\section{Discussion}
\label{sec:discussion-conceptual}

The reaction-time result of Section~\ref{sec:rt-results} should be read conservatively. Its value lies relative to the proposed 60--120~ms window: because the extraction procedure enforces a minimum measurable latency of approximately 46.875~ms, the accumulation of a non-trivial number of events near that boundary before the distribution rises into slower ranges shows that the data contain events at the boundary between fast anticipatory-confirmatory action and more fully elaborated tactical response, with the qualification developed in Section~\ref{sec:latency} that the near-floor cluster itself is better explained by motion already underway than by genuine reactions to the shot. This does not isolate literal limbic discharge, nor does it validate a specific brain structure or a one-to-one \textit{fight-or-flight} mapping. What it validates is more modest: a measurable, interpretable, and highly non-random short-latency response regime with a plausible spread of fast, intermediate, and slow response states, the condition under which \mesocort{} seeks to differentiate rapid reactive control from slower tactical and strategic behavior.

Several caveats bound this interpretation. The earliest bins may reflect a mixture of anticipatory readiness, partial pre-orientation, predictive coding, and rapid sensory confirmation rather than a single mechanism; the slower bins may contain genuine freezing or hesitation alongside residual confounds not yet removed by filtering. The right-skewed shape itself is not diagnostic, since skewed and ex-Gaussian-like distributions are common across many reaction-time tasks~\cite{ratcliff1979,ratcliff2008,luce1986}; we therefore do not treat the histogram shape as evidence of a unique multi-regime mechanism, and stronger claims would require both a control distribution from expected-contact events and formal model comparison using mixture or diffusion-style analyses.

The broader contribution of the current phase is therefore primarily methodological. We show that it is feasible to transform omniscient replay data into a player-centric representation, extract interpretable reaction-centered events, and train predictive models that capture regularities in short-horizon behavior. The neuroscience reviewed here informed the architectural choices (signal selection, model structure, labeling distinctions), but their validation remains empirical: whether the resulting architecture predicts human behavior under pressure. This does not constitute direct measurement of limbic or subcortical processes, nor does it justify strong one-to-one biological claims; it does show that ecologically rich digital environments support measurable, scalable study of pressure-sensitive behavioral organization. Recent EEG findings in esports, showing that the neural state before a match carries predictive information about competitive outcomes, are consistent with this, although our pipeline relies on behavior alone~\cite{minami2023,minami2024}.

If the model captures real structure in how humans behave under sudden threat, especially under out-of-player generalization, then applying it to a new domain becomes primarily a data and validation problem rather than a claim that neural pathways are identical across domains. The same modeling structure (ego-centric information reconstruction, threat-imminence-structured control-mode inference, and a multi-timescale behavioral hierarchy) could in principle be trained on telemetry from driving simulators, surgical training systems, or VR platforms. The transfer question is therefore whether the architecture learns predictive behavioral structure from new data: neuroscience informs the architecture, predictions validate it.

One inferential challenge must be stated explicitly. The claim that fast behavioral responses in CS2 reflect ``limbic-like'' processing is a form of \emph{reverse inference}: reasoning from an observed behavior back to a putative neural mechanism~\cite{poldrack2006,poldrack2011}. Poldrack showed that such inferences are valid only when the neural structure is selectively engaged by the task, which behavioral data alone cannot establish. No pattern in gameplay telemetry can yet distinguish limbic-like processing from fast cortical processing, trained automaticity, anticipatory motor preparation, or strategic pre-orientation without concurrent physiological evidence. The neuroscience vocabulary in this paper should therefore be read as a hypothesis-generating framework rather than as direct evidence of a mechanism.

What the behavioral data can establish is the temporal and structural organization of pressure-sensitive responses (when they occur, how they vary across individuals, and what contextual factors predict them), the prerequisite for future studies pairing the \mesocort{} pipeline with concurrent EEG, pupillometry, eye tracking, or electrodermal measurement. \mesocort{}'s broader significance is thus to make the ``limbic-like'' hypothesis computationally testable: a robust expectation model could identify atypical responses and learn player-specific response profiles under uncertainty.

\section{Limitations}
\label{sec:limitations}

\subsection{Latencies in Counter-Strike~2}
\label{sec:latency}

Reaction latencies are measured from server-side demo files, yet the behavior they describe was produced by a player acting on a \emph{client-side} view of the match. The server records the game state 64 times per second and the demo preserves this record, whereas each client renders a predicted, network-delayed, interpolated reconstruction in which the local player is predicted forward to hide input latency while every other entity is interpolated from earlier server updates, leaving the local player roughly one round-trip ahead of the enemies around them~\cite{bernier2001}. The residual latency measurably degrades player performance~\cite{claypool2006,liu2022survey,xu2022csgo}; the peeker's advantage, governed more by the latency of the player being shot at than by that of the shooter~\cite{tokey2024peekers}, puts a player shot from behind on the disadvantaged side. A demo-measured reaction time therefore contains a purely \emph{technical} delay.

Two delays bracket the player's own reaction time. $\Delta_{\mathrm{stim}}$ is the delay before the player can see the event: the update must reach the client, be rendered, and be displayed. For an enemy coming into view there is an extra interpolation term: the demo stores the enemy at server tick $t_N$, but the screen showed an earlier position (around $t_{N-2}$ to $t_{N-1}$ at 64~Hz), so using $t_N$ as the stimulus time starts the clock before the enemy was visible. For incoming damage, noticed from sound and damage feedback, this term does not apply. $\Delta_{\mathrm{resp}}$ begins after the human reaction and covers mouse polling, the trip back to the server, and rounding to the next recorded tick. Both are strictly positive, so the demo always reports a reaction time longer than the real one: $t_{\mathrm{measured}} \approx t_{\mathrm{true}} + \Delta_{\mathrm{stim}} + \Delta_{\mathrm{resp}}$.

\begin{table}[t]
\caption{Estimated CS2 system-latency contribution to demo-measured reaction time
(typical European competitive conditions). Values are order-of-magnitude estimates,
not measurements.}
\label{tab:latency-budget}
\centering
\begin{tabular}{lc}
\toprule
\textbf{Component} & \textbf{Est. (ms)} \\
\midrule
\multicolumn{2}{l}{\emph{$\Delta_{\mathrm{stim}}$: server $\rightarrow$ player}} \\
\quad Server $\rightarrow$ client (one-way) & 12--18 \\
\quad Client processing + render pipeline (200--400~FPS) & 6--15 \\
\quad Display (144--240~Hz scan-out + pixel response) & 5--12 \\
\quad Interpolation (enemy-in-view trigger only) & 16--31 \\
\midrule
\multicolumn{2}{l}{\emph{$\Delta_{\mathrm{resp}}$: player $\rightarrow$ demo}} \\
\quad Mouse polling + client $\rightarrow$ server (one-way) & 13--23 \\
\quad Quantization to the recording tick & 0--16 \\
\midrule
\textbf{Total delay, shot from behind} & $\approx$\,\textbf{40--80} \\
\quad (stimulus without interpolation $+$ response) & \\
\textbf{Total delay, enemy-in-view} & $\approx$\,\textbf{60--110} \\
\quad (stimulus with interpolation $+$ response) & \\
\bottomrule
\end{tabular}

\vspace{4pt}
\footnotesize\emph{Note:} $\Delta_{\mathrm{stim}}$ is the delay before the player can
perceive an event the server has recorded; $\Delta_{\mathrm{resp}}$ is the delay before
the player's reaction is recorded back into the demo. Network terms vary widely across
players with connection quality (e.g.\ wired versus wireless) and distance to the server.
\end{table}

Table~\ref{tab:latency-budget} gives order-of-magnitude estimates for typical European competitive conditions (a few hundred frames per second, a 144--240~Hz monitor): roughly 40--80~ms when the trigger is incoming damage and 60--110~ms for an enemy coming into view, the difference being the interpolation term. Sub-tick input timing does not help: the server timestamps inputs to a fraction of a millisecond, but the demo is still written once per 64~Hz tick~\cite{valveSubtick}, so the view-angle stream used to find the onset is rounded to the tick. On top of this average offset sits a per-event jitter of tens of milliseconds (tick rounding on both sides, frame-rate changes, drifting ping) that no fixed correction can remove.

These delays change how the back-shot histogram (Fig.~\ref{fig:backshot-hist}) should be read. The large group of reactions in the few-hundred-millisecond range survives the correction. The cluster near the measurement floor ($\approx 46.875$~ms) does not: subtracting a 40--80~ms system delay would put the true reaction time at zero or below, impossible as a response to the shot itself, so these near-floor cases are better explained by movement already underway (anticipation, or leftover mouse motion from an earlier event, Section~\ref{sec:reaction-extraction}) than by genuinely fast reactions. Because the system delay plus jitter is about as large as the 60--120~ms window itself, no single demo-measured latency can show that a reaction fell inside that window. Since the delay grows with ping, which the demo records every tick, filtering to players with low, stable ping would shrink both the average delay and its spread.

\begin{figure}[t]
    \centering
    \includegraphics[width=1.0\linewidth]{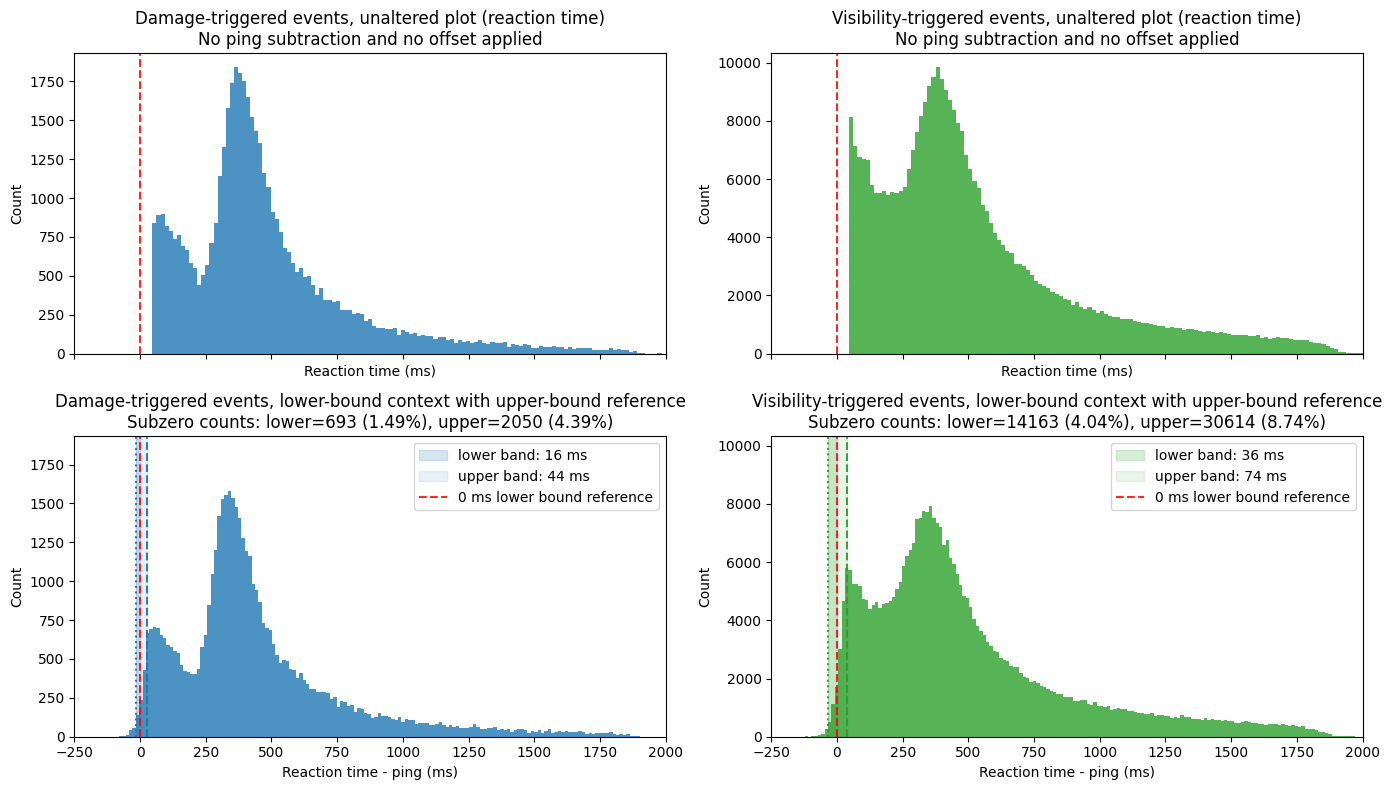}
    \caption{Visualization of reaction time transformation based on estimated unmeasured delay and measured ping. Top row = raw reaction time; bottom row = reaction time minus ping and lower band, with shifted upper band. Minimum reaction time filter before adjustment: 46.875 ms (3 ticks)}
    \label{fig:latency_hist}
\end{figure}

Fig.~\ref{fig:latency_hist} shows the histogram after subtracting the measured ping and the estimated delay bands, exposing the incorrectly included data points. The bands differ from Table~\ref{tab:latency-budget} because each player's exact ping replaces the approximated value. The distribution is not significantly changed, but the case for improved visibility-triggered event filtering arises again. It is purposefully built from all collected reactions rather than the more heavily filtered model dataset, so the two differ in sample size.

\subsection{Statement of Limitations}

None of the following invalidates the findings, but they shape their interpretation.

\textbf{Few-shot example budget.} For cost reasons the commercial LLMs were mostly run with one eighth of the few-shot examples that fit their context window; the single run at three quarters of the budget (Gemini 3.7 Flash, Full-Dump) did not improve accuracy, but the LLM figures should be read as a bounded rather than exhaustive search over prompting configurations. The paired comparisons are on a 1,000-row sample of the validation set, not the full set. The \textit{Jev} comparison is bounded by the model's context limit (at most 120 in-context examples) and is not included in the paired tests.

\textbf{Model selection and held-out evaluation.} Models were selected on validation loss and the headline accuracies are reported on the same validation set; no separate held-out test set was reserved for the main experiments (the player-disjoint experiments are the exception), so the reported figures should be read as an upper bound on generalization.

\textbf{Interval estimates.} The headline accuracies are point estimates without confidence intervals, and no retraining variance across seeds is reported; paired significance tests are given for the headline comparisons only, so differences of a few percentage points between LLM configurations in Table~\ref{tab:cross_model_comparison_table} should not be treated as established.

\textbf{Independence of observations.} Reaction moments are nested within players and matches, whereas McNemar's test treats the 1,000 sampled rows as independent. The reported $p$-values may therefore overstate the evidence to an extent that has not been quantified; a test clustered by player and match is left to future work.

\textbf{Label reliability.} Inter-annotator agreement is Fleiss's kappa $\kappa = 0.38$ (200 moments) and $0.44$ (gold set), fair and moderate on the Landis and Koch scale. This bounds the accuracy achievable by any model trained against these labels, so the reported accuracies should be read against that ceiling rather than against 100\%; part of the residual error may reflect genuine ambiguity.

\textbf{Consensus coverage.} The 82.5\% consensus coverage is the proportion of moments on which a majority was reached. With seven annotators and a binary choice, a majority is near-certain unless annotators abstain, so it principally reflects abstention and is not an agreement statistic; $\kappa$ is.

\textbf{Label circularity.} \textit{Fight-or-flight} labels are coarse operational approximations derived from the same telemetry stream as the features, so classifier performance partly reflects internal consistency. The gold-standard check (80.6\% agreement) mitigates but does not remove this, and the gold set covers damage events only.

\textbf{Temporal resolution and system latency.} The nominal 64~Hz tick (15.625~ms) bounds the resolution of every measurement. Measured latencies are upper bounds on true reaction times, inflated by roughly 40--80~ms of system delay plus per-event jitter of tens of milliseconds (Section~\ref{sec:latency}), and the 46.875~ms floor introduces left-censoring that may inflate apparent clustering at short latencies.

\textbf{Sequence-history behavior.} The degradation with longer history windows in Table~\ref{tab:orbit_history_window_results} is not expected theoretically and is most likely a training artifact; no conclusion depends on it.

\textbf{Absent baselines and controls.} The back-shot histogram lacks a control distribution from non-surprise contacts, preventing claims of threat-specificity, and its unimodal versus multimodal explanations are not formally tested by mixture or diffusion-style modeling~\cite{ratcliff1979,ratcliff2008,luce1986}. Furthermore, ORBIT has not been compared against feature-ablated naive baselines, limiting how far its loss curves evidence generalization.

\textbf{Scope.} The data come from a single game, a single heavily filtered scenario, and demos selected around a core cohort of 50 players, and may not transfer to other environments without revalidation. Auditory information is not yet fully reconstructed, and the Ego-Knowledge layer has not been validated against perception-oriented ground truth such as eye tracking. There is no physiological ground truth, so references to limbic-like behavior remain inferential (reverse inference) until concurrent neuroimaging or psychophysiological measurement is available~\cite{poldrack2006}. Neuroimaging of experienced FPS players suggests tonic amygdala suppression during sustained gameplay~\cite{montag2012,weber2006}, which constrains the expected signal to transient event-locked responses. Event extraction may also be confounded by skill, role, map knowledge, automaticity, or contextual priors not yet represented explicitly~\cite{fitts1967}.

\textbf{Latency bands and mechanism.} Reaction-time bands are named by latency; the mapping from latency to control mode is a hypothesis under test, and no claim is made that a band identifies a mechanism.

\textbf{Architecture.} Hand-engineered features and behavioral cloning are substantially behind the frontier of world-model architectures; the roadmap in Section~\ref{sec:future-work} addresses this through learned encoders and action-conditioned prediction.

\section{Ethics and Data Protection}\label{sec:ethics}

\textbf{Nature of the data.} This work is a secondary analysis of recorded gameplay telemetry from Counter-Strike~2 matches, extracted from server-side GOTV demo files. No participants were recruited and no intervention took place. The demo files were obtained from two sources. The primary source is the Rankacy platform, where users upload or record their matches and, under the platform's Terms of Service and Privacy Policy, consent to the processing of their gameplay data, including derived analysis and use for research and scientific purposes. A second portion of the dataset was purchased from third-party operators who ran their own game servers and generated the demo files on their own infrastructure; these datasets were acquired under contractual terms permitting retention, derived analysis, and publication of derived, pseudonymized results. No demo files were scraped from third-party services without authorization.

\textbf{Controller and legal basis.} The data controller is JMC a.s. (Czech Republic), which operates the Rankacy platform and conducts the research described here under the Mesocort AI name; the applicable terms are published at rankacy.com. Processing for this research relies on Article 6(1)(f) GDPR (legitimate interests), documented in a Legitimate Interests Assessment, alongside the contractual and consent framework of the platform's Terms of Service, and is carried out with the safeguards for scientific research under Article 89(1) GDPR, principally pseudonymization and data minimization. A Data Protection Impact Assessment under Article 35 GDPR covering behavioral profiling research on gameplay telemetry has been carried out and is maintained by the controller.

\textbf{Identifiers and pseudonymization.} Player identifiers are replaced by sequential labels (player~1, player~2, \ldots) in all published material and in any released dataset. Identifiers are never perturbed, because platform identifiers occupy a contiguous number space in which an altered identifier resolves to a different real account. The identifier-to-label mapping is retained by the controller under restricted access, solely to support scientific validation, correction, and the exercise of data-subject rights; the data are therefore treated as pseudonymized personal data and the GDPR applies in full. Longitudinal behavioral traces also carry a residual re-identification risk, which is a further reason the data are described as pseudonymized rather than anonymous.

\textbf{Ethics review.} No institutional ethics committee review was sought for this work. The study is a secondary analysis of previously recorded gameplay telemetry processed under the contractual and data-protection framework described above; no participants were recruited, no intervention or interaction with individuals took place, and no special categories of personal data under Article 9 GDPR were processed. The research was conducted by a commercial entity outside a mandatory institutional review framework, and oversight was provided through the controller's internal data-protection process. Future phases involving physiological measurement (EEG, eye tracking) will be conducted with academic partners under their institutional ethics approval.

\textbf{Transparency.} The individuals whose gameplay appears in the dataset were not contacted individually about this study. Users of the Rankacy platform were informed through the platform's Terms of Service and Privacy Policy at the point of collection that their data may be processed for research and scientific purposes (Article 13 GDPR). For records originating from purchased third-party datasets, individual notification would require disproportionate effort within the meaning of Article 14(5)(b) GDPR; in line with that provision, the safeguards of Article 89(1) are applied, and this publication, together with the information available at rankacy.com, serves as the public information measure.

\textbf{Intended use and misuse.} Because this line of research concerns behavior prediction and behavioral profiling, it requires explicit safeguards around privacy, data governance, fairness, uncertainty reporting, and misuse prevention. A system that models pressure-sensitive or player-specific response tendencies may be valuable for training and simulation, but it also carries risks if used for manipulative personalization, covert profiling, or surveillance. The models described here are developed for research purposes and are not intended for individual-level profiling of identifiable persons.

\section{Conclusion}
This work positions Mesocort as an effort to model a fast, pressure-sensitive layer of human behavior from high-frequency digital traces. The strongest near-term claim is not that the system reads the limbic system directly, but that it can detect and organize behavioral proxies of rapid, high-pressure responses; the neuroscience reviewed here informed the architecture, and its validation remains empirical.

The two evaluation regimes are complementary. Under strictly pre-reaction information, the \textit{fight-or-flight} outcome is predictable well above the $51.5\%$ majority-class baseline: EMBER reaches $73.5\%$ and ORBIT $71.2\%$ on the full validation set, and both generalize to held-out players. The strongest LLM baseline reaches $62.7\%$ on 1,000 predictions from the same validation set; on those same trials EMBER ($73.6\%$) and ORBIT ($70.9\%$) outperform it with paired McNemar $p = 1.3\times10^{-9}$ and $p = 1.5\times10^{-6}$, respectively, and the best LLM accuracy did not improve across a generation of newer models. Both telemetry-native models predict at several orders of magnitude lower inference time and cost, although that comparison is not compute-matched. In Early Response Recognition, accuracy rises monotonically with added post-anchor ticks, reaching $85.15\%$ over the full ${\approx}1$\,s window. Part of this increase is mechanical, since the label is defined by post-anchor behavior; the informative quantity is the \emph{shape} of the curve, along which a fast reactive component is expected to dominate early and a slower, deliberative one to assert itself later, sometimes revising the early intention. Marginal gains per added tick are roughly uniform, so this remains a hypothesis rather than a result the data establish.

These results were obtained on a deliberately small slice of the available data (demos selected around a $50$-player core cohort and one heavily filtered surprise scenario) and are a Phase~1 proof of feasibility rather than a converged figure. The two models perform comparably, with EMBER slightly ahead: leakage minimization leaves little temporal signal at the anchor, and ORBIT overfits well before the end of training (Figure~\ref{fig:orbit_7tick_loss}), a symptom of data-hungry attention-based models in this low-data regime \cite{dosovitskiy2021image,kaplan2020scaling}. We nonetheless prioritize the transformer direction deliberately: the goal is a single model spanning many behavioral use cases, where shared representations and transfer matter more than single-task accuracy, and that direction depends on larger scale and richer temporal inputs, under which attention-based models are expected to gain. ORBIT's stable training is read as a base for those settings rather than present superiority; EMBER remains a strong, low-cost model where temporal context is uninformative.

Mesocort complements environment world models and language models by targeting a different layer: not how environments evolve, nor deliberative reasoning and communication, but embodied, individual-specific, pressure-sensitive dynamics learned from telemetry. The LLM comparison, now paired on identical trials, is consistent with native telemetry models capturing variance in the fast-response regime that text-based general-purpose models are structurally poorly suited to access, although it is not compute-matched. The long-term vision, a validated human behavior prior populating world-model simulations with realistic agents, depends first on proving the prediction under out-of-player generalization and on resolving the reverse-inference constraint through physiological measurement.

\section{Future Work}\label{sec:future-work}
The present study establishes a Phase~1 baseline; the most consequential next steps extend the evidence base, input signal, data scale, and behavioral phenomena captured.

\begin{enumerate}[leftmargin=*, itemsep=4pt]
    \item \textbf{Confidence intervals, held-out test set, and further baselines.} Reporting confidence intervals for every headline figure, reserving a held-out test set separate from the validation set used for model selection, and adding a pretrained non-LLM tabular baseline such as TabPFN together with a compute-matched, locally served LLM comparison and a full, paired evaluation of decision models such as \textit{Jev}.

    \item \textbf{Richer sensory reconstruction (auditory channel).} Phase~1 reconstructs vision but not sound. Footsteps, gunfire, reloads, and grenade audio are primary \emph{anticipatory} cues in Counter-Strike~2, and their absence confounds reaction-onset measurement, since a visually attributed response may have been sound-cued earlier. An auditory layer would remove this confound and open the study of anticipatory behavior.

    \item \textbf{Scaling the dataset.} Results rest on a constrained set of heavily filtered surprise events, limiting statistical power for control distributions from non-surprise contacts, player-disjoint generalization, and rare edge cases such as freezing. Larger coverage is a prerequisite for inferential rigor and the foundation-model direction below, and must include the gold-standard subset.

    \item \textbf{Reducing the system-latency confound.} Reaction times read from demo files are upper bounds on true latency, inflated by a strictly positive engine-and-network delay of roughly $40$--$80$\,ms for back-damage events (Section~\ref{sec:latency}). Because the demo records each player's ping every tick, filtering to low, stable connections would shrink the average offset and its per-event jitter without new data.

    \item \textbf{Strengthening the ORBIT evaluation protocol.} ORBIT results do not yet report non-learned baselines beyond the majority class, nor a fully documented strict-split strategy. Future iterations will add naive and feature-ablated baselines and an explicitly documented player- and match-disjoint splitting protocol.

    \item \textbf{A telemetry-native foundation model for Counter-Strike~2.} We intend to pretrain a self-supervised foundation model on unlabeled CS2 telemetry at scale, with a learned agent-centric encoder replacing hand-engineered features and the visibility engine as a training-time leak filter, serving as a shared backbone for downstream tasks and as the substrate for upgrading ORBIT from behavioral cloning to an action-conditioned, counterfactual agent world model.

    \item \textbf{Modeling the transition from fast to deliberative control.} Dual-route accounts described a subcortical ``low road'' preceding a cortical ``high road''~\cite{ledoux1996,ledoux2000}, but that binary has since been superseded: LeDoux and Daw recast defensive behavior as a hierarchy of controllers within a single survival circuit~\cite{ledoux2018}, and Pessoa and Adolphs argue for distributed relevance evaluation rather than a privileged shortcut~\cite{pessoa2010}. We therefore name our bands by latency rather than by pathway, and treat the mapping from latency to control mode as the hypothesis under test. The target is the handoff between the fast and slower bands, modeled explicitly via, for instance, a switching-state-space formulation gated by threat-imminence features. A first test needs no new recordings: retrain with the fastest corrected-latency band removed from the training set, compare against removing an equal number of slower events, and measure the change on the full validation set, starting with damage-triggered events, where the stimulus anchor is cleanest. If accuracy does not move, the fast band carries no predictive signal of its own and the fast-response framing should be narrowed accordingly.

    \item \textbf{Validation, physiology, and transfer.} In parallel we will pursue a larger gold-standard set to reduce label circularity, formal statistical treatment of the reaction-time distribution (control conditions, mixture- and diffusion-style fits), synchronized physiological measurement (EEG, pupillometry, electrodermal) where partnerships allow, to address the reverse-inference constraint, and architectural transfer to at least one non-gaming pressure domain.
\end{enumerate}

\section{Data and Code Availability}
The trained models (EMBER and ORBIT), the extracted model-input data underlying the reported results, and the evaluation protocol are available on reasonable request to qualified research groups for validation and replication purposes, subject to a research data-sharing agreement preserving pseudonymization and prohibiting re-identification. The raw demo corpus, the parsing and Ego-Knowledge reconstruction pipeline, and the visibility engine are proprietary and are not released, as they form part of Mesocort AI's commercial platform. Data are otherwise available from the corresponding author on reasonable request.

\section{Competing Interests}
Several authors are employees, contractors, or shareholders of Mesocort AI, which operates within the JMC a.s. group and is developing commercial applications of the models described in this paper. The authors declare no other competing interests.

\section{Funding}
This research was funded entirely by private capital of the JMC a.s. group; no public grants or external funding were used.

\section{Acknowledgments}
We thank Dr. Marcel Binz for his thorough review of this work and for his valuable comments and suggestions.

\onecolumn

\appendix

\subsection{Input features used by the ORBIT model in detail}
\label{sec:appendix-input-features}
\begingroup
\small
\setlength{\LTleft}{0pt}
\setlength{\LTright}{0pt}
\setlength{\tabcolsep}{4pt}
\renewcommand{\arraystretch}{1.08}

\begin{longtable}{L{0.22\textwidth} L{0.56\textwidth} L{0.18\textwidth}}
\caption{Description of the input features used by the model. All features are related for the current game tick.}
\label{tab:dataset_features_all} \\
\toprule
\textbf{Feature} & \textbf{Description} & \textbf{Encoding / scaling} \\
\midrule
\endfirsthead

\multicolumn{3}{c}%
{{\tablename\ \thetable{} -- continued from previous page}} \\
\toprule
\textbf{Feature} & \textbf{Description} & \textbf{Encoding / scaling} \\
\midrule
\endhead

\midrule
\multicolumn{3}{r}{\textit{Continued on the next page}} \\
\endfoot

\bottomrule
\endlastfoot

\multicolumn{3}{l}{\textbf{Round and map context}} \\ \midrule Premier rank & Player Premier rank. & Min--max scaler \\ Round number & Current round index. & Embedding layer \\ Map ID & Identifier of the map being played. & Embedding layer \\ Bomb planted & Indicates whether the bomb is currently planted. & Boolean \\ Remaining round time & Remaining time in the current round. & Standard scaler \\ \midrule \multicolumn{3}{l}{\textbf{Player position and state}} \\ \midrule Player square ID & Discretized player position on the map. The map space is divided into squares, each represented by a pretrained embedding. & Embedding layer \\ Square local X & Player's local X coordinate inside the current square & Min--max scaler \\ Square local Y & Player's local Y coordinate inside the current square. & Min--max scaler \\ Global Z position & Player's vertical position on the map. & Per-map min--max scaler \\ Flash duration & Duration for which the player is blinded by a flashbang. & Standard scaler \\ View offset Z & Height of the player's head or view origin. & Min--max scaler \\ Alive & Indicates whether the player is alive. & Boolean \\ Health & Current health of the player. & Min--max scaler \\ Team & Indicates whether the player belongs to the T or CT team. & Boolean \\ Pitch & Vertical angle of the player's view direction. & Sine transform \\ Yaw cosine & Cosine component of the player's yaw angle. & Cosine transform \\ Yaw sine & Sine component of the player's yaw angle. & Sine transform \\ Weapon ID & Identifier of the currently equipped weapon. & Embedding layer \\ Two-dimensional velocity & Player velocity in the horizontal plane, excluding the Z component. & Min--max scaler \\ Field of view & Player field of view. & Min--max scaler \\ Has defuser & Indicates whether the player carries a defuse kit. & Boolean \\ Has bomb & Indicates whether the player carries the bomb. & Boolean \\ Number of team defusers & Number of defuse kits available in the player's team. & Min--max scaler \\ Defusing time & Time for which the player has been defusing the bomb. & Standard scaler \\ Bomb planting time & Time for which the player has been planting the bomb. & Standard scaler \\ Team has bomb & Indicates whether the player's team currently controls the bomb. & Boolean \\ \midrule \multicolumn{3}{l}{\textbf{Aiming and visibility}} \\ \midrule Angular velocity & Change in the player's view angle between the previous tick and the current tick. This feature captures the speed of mouse or view movement. & Power transformer \\ Aim direction state & Discrete state describing the player's aim direction. & Embedding layer \\ Aim direction changed & Indicates whether the player's crosshair direction changed from the previous tick. & Boolean \\ Closest enemy to crosshair, Euclidean distance & Euclidean distance between the crosshair direction and the closest enemy. & Power transformer \\ Closest enemy angular distance & Angular distance between the crosshair and the closest enemy. & Power transformer \\ Closest enemy Euclidean distance & Euclidean distance to the closest enemy. & Power transformer \\ Number of visible enemies & Number of enemies visible to the player. & Min--max scaler \\ Some enemies visible & Indicates whether at least one enemy is visible. & Boolean \\ Some enemies visible, uncertain & Indicates uncertain enemy visibility, for example due to smoke boundaries or partial occlusion. & Boolean \\ Enemy visible duration & Number of ticks for which an enemy has been continuously visible. & $\log(1+x)$ + standard scaler \\ Ticks since last enemy visible & Number of ticks since an enemy was last visible. & $\log(1+x)$ + standard scaler \\ Has seen enemy in round & Indicates whether the player has seen an enemy at least once in the current round. & Boolean \\ Player in smoke & Indicates whether the player is currently inside smoke. & Boolean \\ \midrule \multicolumn{3}{l}{\textbf{Player input actions}} \\ \midrule Movement and action buttons & Indicates whether the player currently holds one of the following input buttons: jump, fire, use, crouch, walk, forward, back, left, or right. & Boolean \\ Weapon fired & Indicates whether the player fired a weapon in the current tick. & Boolean \\ Ticks since last weapon fire & Number of ticks since the player last fired a weapon. & $\log(1+x)$ + standard scaler \\ Has fired weapon in round & Indicates whether the player has fired a weapon in the current round. & Boolean \\ \midrule \multicolumn{3}{l}{\textbf{Damage and kill events}} \\ \midrule Damage given & Indicates whether the player dealt damage in the current tick. & Boolean \\ Ticks since last damage given & Number of ticks since the player last dealt damage. & $\log(1+x)$ + standard scaler \\ Has given damage in round & Indicates whether the player has dealt damage in the current round. & Boolean \\ Damage received & Indicates whether the player received damage in the current tick. & Boolean \\ Ticks since last damage received & Number of ticks since the player last received damage. & $\log(1+x)$ + standard scaler \\ Has received damage in round & Indicates whether the player has received damage in the current round. & Boolean \\ Team suffered kill & Indicates whether the player's team lost a player in the current tick. & Boolean \\ Ticks since last team suffered kill & Number of ticks since the player's team last lost a player. & $\log(1+x)$ + standard scaler \\ Has team suffered kill in round & Indicates whether the player's team has lost at least one player in the current round. & Boolean \\ Enemy team suffered kill & Indicates whether the enemy team lost a player in the current tick. & Boolean \\ Ticks since last enemy team suffered kill & Number of ticks since the enemy team last lost a player. & $\log(1+x)$ + standard scaler \\ Has enemy team suffered kill in round & Indicates whether the enemy team has lost at least one player in the current round. & Boolean \\ Kill given & Indicates whether the player killed an enemy in the current tick. & Boolean \\ Has given kill in round & Indicates whether the player has killed an enemy in the current round. & Boolean \\ \midrule \multicolumn{3}{l}{\textbf{Teammate features}} \\ \midrule Teammate $i$ distance & Distance between the player and teammate $i$, where $i \in \{1,\ldots,4\}$. & Standard scaler \\ Teammate $i$ visible & Indicates whether teammate $i$ is visible to the player. & Boolean \\ Teammate $i$ square ID & Discretized map square occupied by teammate $i$. & Embedding layer \\ Teammate $i$ square local X & Local X coordinate of teammate $i$ inside the corresponding map square. & Min--max scaler \\ Teammate $i$ square local Y & Local Y coordinate of teammate $i$ inside the corresponding map square. & Min--max scaler \\ Teammate $i$ global Z position & Vertical position of teammate $i$ on the map. & Per-map min--max scaler \\ \midrule \multicolumn{3}{l}{\textbf{Enemy features}} \\ \midrule Enemy $i$ alive & Indicates whether enemy $i$ is alive, where $i \in \{1,\ldots,5\}$. & Boolean \\ Enemy $i$ visible & Indicates whether enemy $i$ is visible to the player. & Boolean \\ Enemy $i$ can see player & Indicates whether enemy $i$ has line of sight to the current player. & Boolean \\ Enemy $i$ square ID & Discretized map square occupied by enemy $i$. & Embedding layer \\ Enemy $i$ square local X & Local X coordinate of enemy $i$ inside the corresponding map square. & Min--max scaler \\ Enemy $i$ square local Y & Local Y coordinate of enemy $i$ inside the corresponding map square. & Min--max scaler \\ Enemy $i$ global Z position & Vertical position of enemy $i$ on the map. & Per-map min--max scaler \\ Enemy $i$ distance & Distance between the player and enemy $i$. & Standard scaler \\ Enemy $i$ crouching & Indicates whether enemy $i$ is crouching. & Boolean \\ Enemy $i$ weapon fired & Indicates whether enemy $i$ fired a weapon in the current tick. & Boolean \\ Enemy $i$ visibility ratio & Fraction of the enemy model visible to the player. The value decreases when the enemy is partially occluded by an obstacle. & Min--max scaler \\ Enemy $i$ crosshair distance & Distance between the player's crosshair and enemy $i$. & Power transformer \\ Enemy $i$ source radial velocity & Radial velocity of enemy $i$ relative to the player. & Standard scaler \\ \end{longtable}
\endgroup

Table \ref{tab:dataset_features_all} describes the input of the transformer model (ORBIT). The dataset used by the gradient-boosted-tree architecture (EMBER) is very similar; embedding and scaling are not used for EMBER, and the remaining differences are listed below:

\begin{itemize}
    \item player, teammate and enemy position is described by standard X, Y, Z values
    \item view offset Z is not used
    \item alive column is not used
    \item field of view is not used
    \item aim direction state and changed are not used
    \item some enemies visible and has seen enemy in round are not used
    \item movement and action buttons are not used
    \item damage given and received features are not used
    \item number of enemies and teammates alive are added
    \item closest teammate distance is added
\end{itemize}

\clearpage

\subsection{Complete Feature-Importance Rankings}

\begin{figure}[H]
    \centering
    \includegraphics[width=\textwidth]{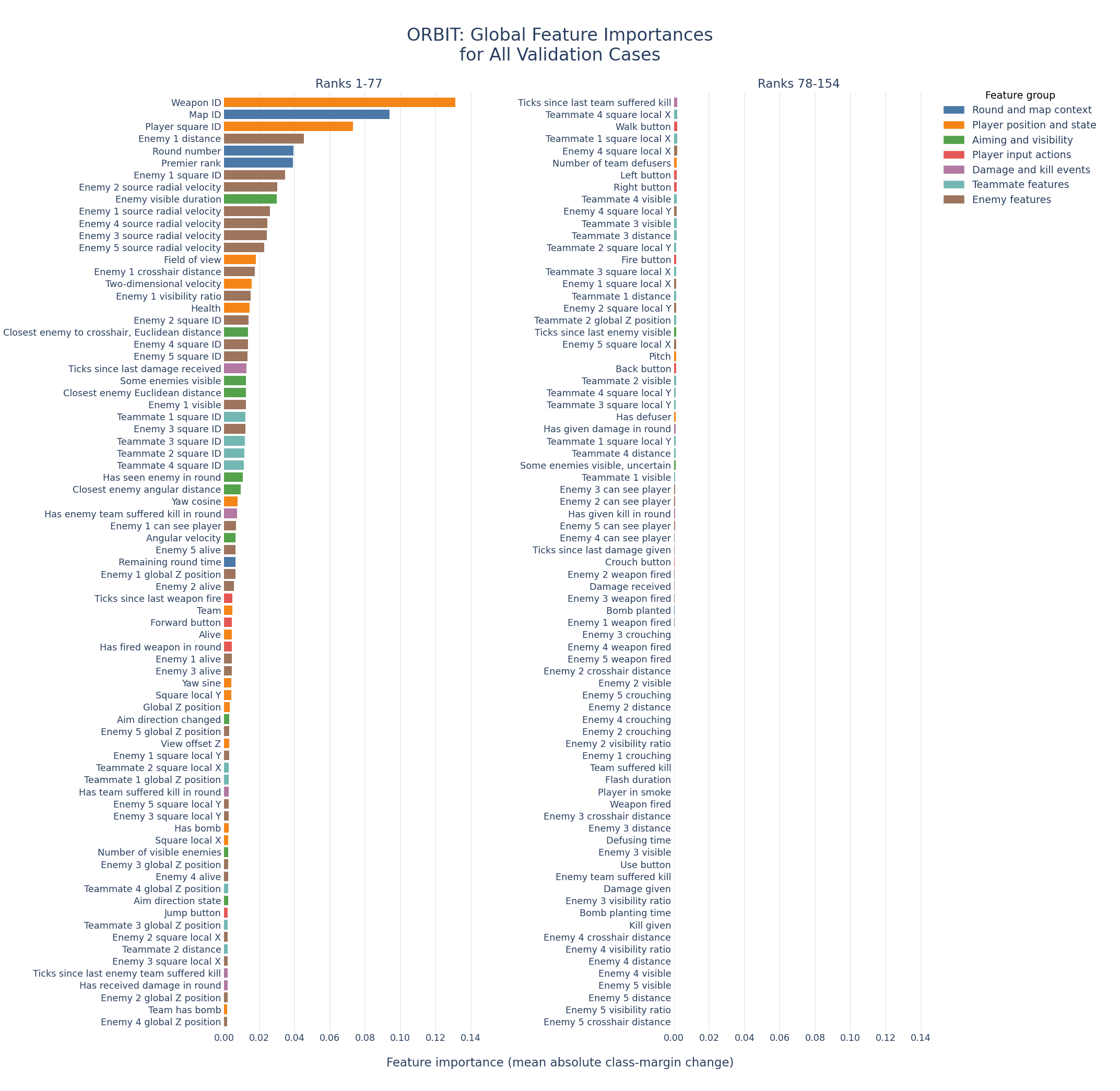}
    \caption{Complete ORBIT global feature-importance ranking. Feature importance is measured as the mean absolute change in the predicted \textit{fight} or \textit{flight} class margin after replacing each feature with a neutral baseline value. Features are ordered by decreasing importance.}
    \label{fig:orbit-global-feature-importance-fight-all-features}
\end{figure}

\begin{figure}[H]
    \centering
    \includegraphics[width=\textwidth]{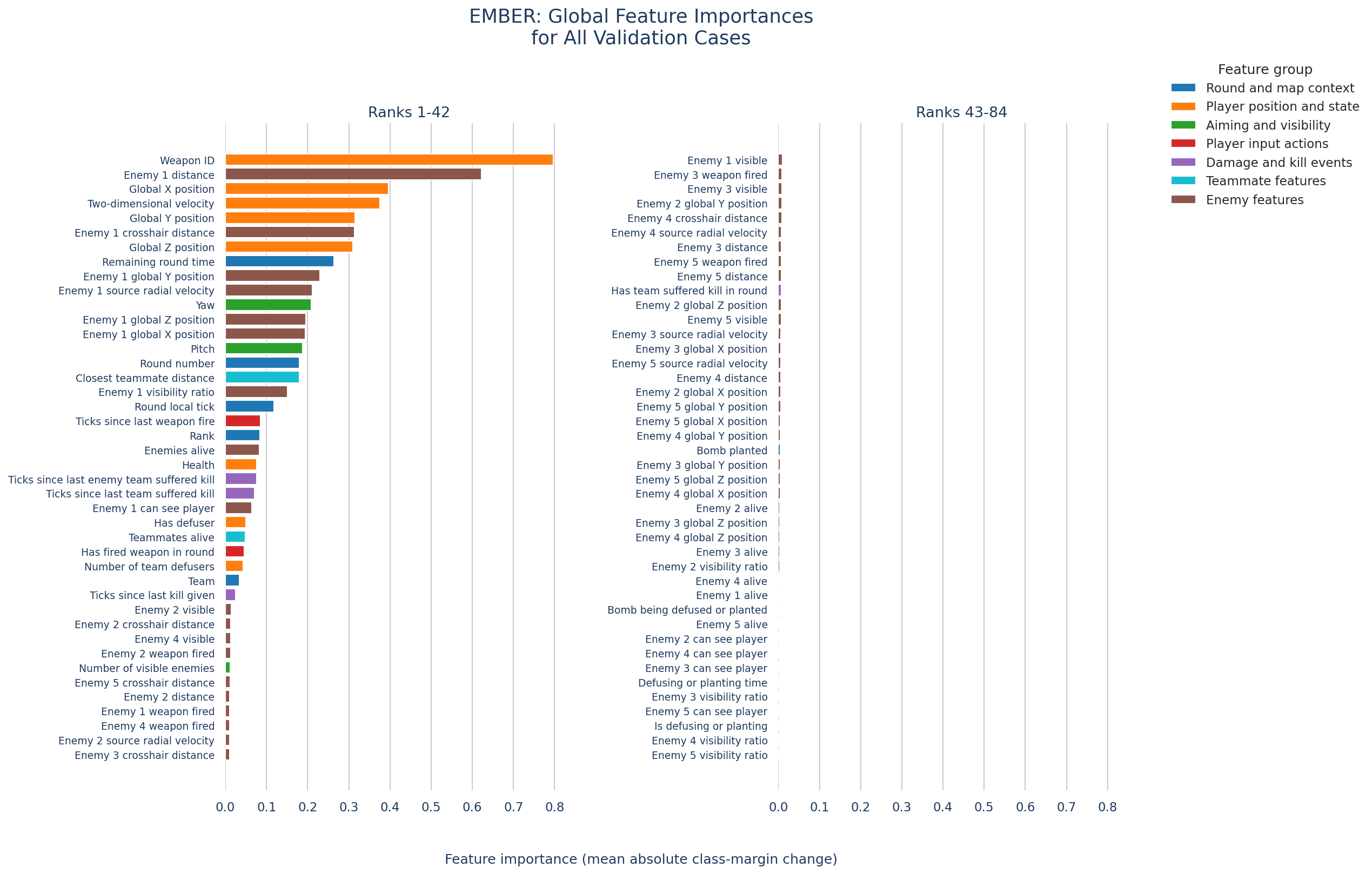}
    \caption{Complete global feature-importance ranking for EMBER. Sorted by gain. EMBER uses a subset of previously laid out feature set.}
    \label{fig:xgb-feats-1}
\end{figure}

\clearpage

\subsection{Data serialization to LLM input}
\begin{figure}[h]
  \centering
  \small
  \begin{tikzpicture}[
    node distance=6mm and 8mm,
    box/.style={
      draw, rounded corners=2pt, align=center,
      minimum width=28mm, minimum height=8mm, inner sep=3pt
    },
    databox/.style={
      box, fill=black!4, minimum width=34mm
    },
    encbox/.style={
      box, fill=blue!6, minimum width=30mm
    },
    promptbox/.style={
      box, fill=orange!8, minimum width=34mm
    },
    apibox/.style={
      box, fill=green!8, minimum width=26mm
    },
    pre/.style={
      box, fill=gray!10, minimum width=34mm, minimum height=10mm
    },
    arrow/.style={-{Stealth[length=2.2mm]}, thick},
    dashedarrow/.style={arrow, dashed},
    grouplabel/.style={
      font=\footnotesize\itshape, text=black!65
    }
  ]
  \node[databox] (parquet) {%
    \textbf{Parquet row}\\[-1pt]
    \scriptsize demo, dt, tick, steamid, label\\
    \scriptsize + flat feature columns
  };
  \node[pre, below=of parquet] (pre) {%
    \textbf{Row preparation}\\[-1pt]
    \scriptsize Drop metadata: demo\_id, dt, tick, steamid\\
    \scriptsize Describe: item\_def\_idx $\rightarrow$ weapon alias\\
    \scriptsize Sanitize: NaN/Inf $\rightarrow$ null (JSON paths)\\
    \scriptsize Keep $-1$ as missing sentinel
  };
  \draw[arrow] (parquet) -- (pre);
  \node[encbox, below left=14mm and 22mm of pre] (csv) {%
    \textbf{Full CSV}\\[-1pt]
    \scriptsize All feature columns\\
    \scriptsize Header + rows
  };
  \node[encbox, below=14mm of pre] (compact) {%
    \textbf{Compact top-$K$ JSON}\\[-1pt]
    \scriptsize Subset from feature importance\\
    \scriptsize Nest: player / enemy\_$i$ / metric / $t$
  };
  \node[encbox, below right=14mm and 22mm of pre] (narr) {%
    \textbf{Narrative top-$k$ text}\\[-1pt]
    \scriptsize Subset from feature importance\\
    \scriptsize Rule-based phrases
  };
  \draw[arrow] (pre.south) -- ++(0,-4mm) -| (csv.north);
  \draw[arrow] (pre.south) -- (compact.north);
  \draw[arrow] (pre.south) -- ++(0,-4mm) -| (narr.north);
  \node[promptbox, below=22mm of compact] (prompt) {%
    \textbf{Prompt assembly}\\[-1pt]
    \scriptsize Task preamble (\textit{fight/flight}, time keys, $-1$ rule)\\
    \scriptsize Few-shot: class-balanced train examples\\
    \scriptsize Query: one validation row (unlabeled)\\
    \scriptsize Output spec: JSON \texttt{\{label, reasoning\}}
  };
  \draw[arrow] (csv.south) -- ++(0,-3mm) -| ($(prompt.north west)!0.15!(prompt.north)$);
  \draw[arrow] (compact.south) -- (prompt.north);
  \draw[arrow] (narr.south) -- ++(0,-3mm) -| ($(prompt.north east)!0.15!(prompt.north)$);
  \node[apibox, below=of prompt] (api) {%
    \textbf{LLM request}\\[-1pt]
    \scriptsize OpenAI / OpenRouter / Gemini / Ollama\\
    \scriptsize Chat: system prefix + user query\\
    \scriptsize or single cached content block
  };
  \draw[arrow] (prompt) -- (api);
  \begin{scope}[on background layer]
    \node[grouplabel, anchor=south west] at ($(csv.north west)+(60.5mm,9.5mm)$) {Structured~~encodings};
  \end{scope}
  \end{tikzpicture}
  \caption{%
    General serialization pipeline from the source dataset row to LLM input.
    All benchmarks share row preparation; encodings differ in how features are rendered.
    All runs attach labeled train examples and one unlabeled query;
    the model returns structured JSON with \texttt{label} and \texttt{reasoning}.%
  }
  \label{fig:llm-serialization-pipeline}
\end{figure}
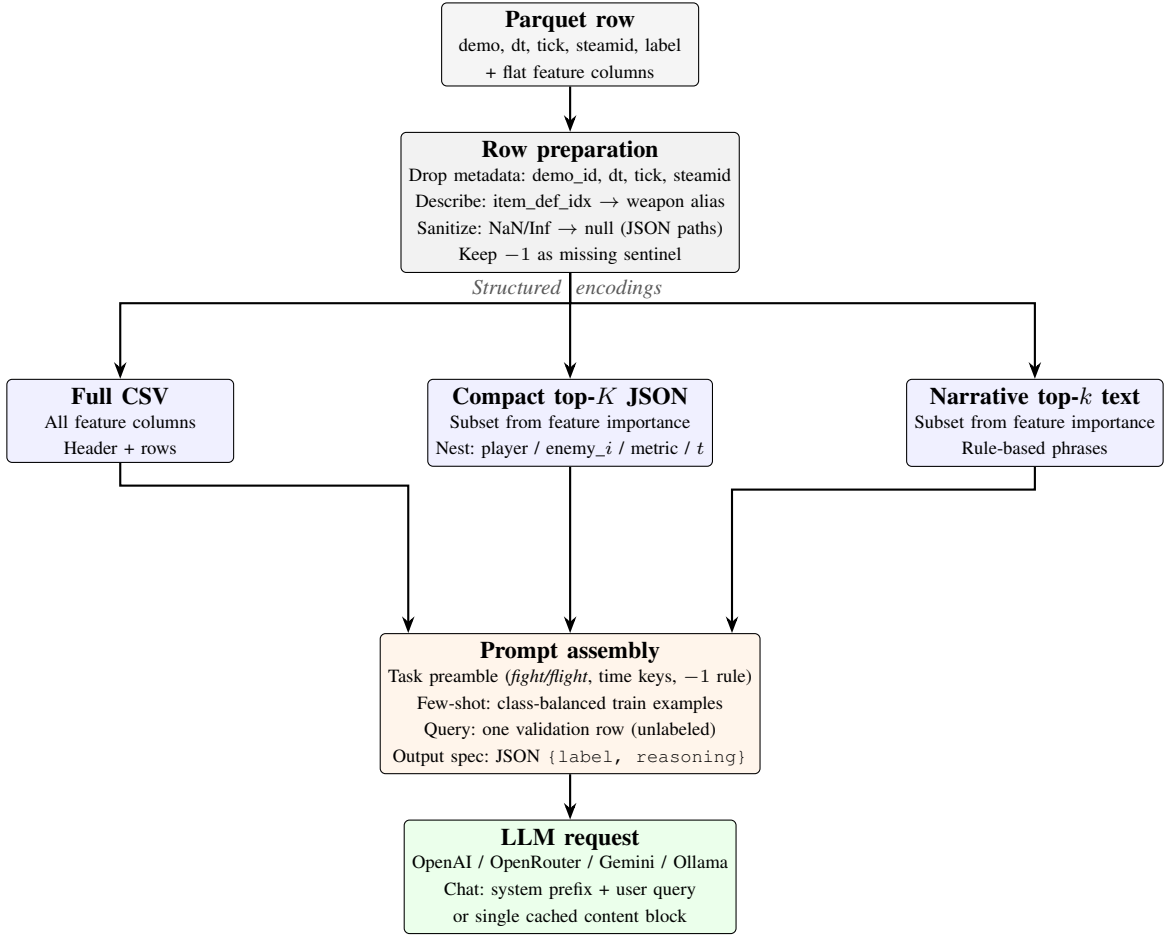

\subsection{LLM prompt examples}
\label{sec:appendix-llm-prompt-examples}

\begin{lstlisting}[caption=Task preamble and label mapping of the LLM prompt using Narrative encoding (excerpt)]
You are a classifier for Counter-Strike 2 (5v5) situations.

Your task is to predict a binary label from game-state features.

Use this label mapping exactly:
- 0 = flight
- 1 = fight

Meaning:
- flight (0): the player is more likely to disengage, retreat, avoid direct confrontation, reposition defensively, or delay contact.
- fight (1): the player is more likely to take or continue engagement, peek, challenge, chase, hold for a duel, or commit to combat.

Constraints:
- Use only the provided feature values.
- Do not assume access to external game knowledge beyond the feature names and values.
- Do not invent missing information.
- Base the decision on the current snapshot and recent history included in the features.

Missing / unknown values:
- Usually -1 or -1.0 means unknown or unavailable information, not a real physical value.
- Treat these values as missing information, not as literal coordinates, distances, or measurements.
\end{lstlisting}

The complete prompt templates for the Narrative, Compact JSON and Full-CSV encodings, including the few-shot examples and the serialized query rows, are available from the authors on request.

\subsection{Recognized patterns in LLM reasoning}
\begin{table}[h]
  \centering
  \caption{\textit{Fight-or-flight} patterns in LLM reasoning}
  \label{tab:fight-flight-patterns}
  \begin{tabularx}{\linewidth}{@{}lX@{}}
    \toprule
    \textbf{Pattern} & \textbf{Description} \\
    \midrule
    \multicolumn{2}{@{}l}{\textbf{Fight patterns}} \\
    \addlinespace[0.25em]
    \textit{The Stationary Hold} &
    Remaining stationary with a primary weapon (especially \textit{AWP}) while the crosshair is aligned with a potential entry point. \\
    \textit{Aggressive Pursuit} &
    Moving quickly toward an enemy or tracking a retreating enemy with a combat-ready weapon. \\
    \textit{Tactical Ambush} &
    Engaging an unaware enemy from a position of advantage, often described as \textit{"stalking"} or \textit{"initiative"}. \\
    \textit{Weapon Switching} &
    Transitioning from a knife or utility to a primary weapon immediately upon spotting an enemy. \\
    \midrule
    \multicolumn{2}{@{}l}{\textbf{Flight patterns}} \\
    \addlinespace[0.25em]
    \textit{Utility/Knife Rotation} &
    Moving at maximum velocity with a knife or grenade out, interpreted as prioritizing speed for repositioning. \\
    \textit{Defensive Repositioning} &
    Moving away from enemy coordinates or turning the view (\texttt{Yaw}) away from the threat to find cover. \\
    \textit{Survival Preservation} &
    Retreating due to critically low health or being the last player alive in a 1vX situation. \\
    \textit{Long-Range Avoidance} &
    Disengaging when the enemy is at extreme distance ($>2000$ units) and the current weapon is ineffective. \\
    \midrule
    \multicolumn{2}{@{}l}{\textbf{Error patterns}} \\
    \addlinespace[0.25em]
    \textit{Crosshair Proximity Bias} &
    Models frequently predict \textit{fight} simply because the crosshair is near an enemy, ignoring that the player is actually retreating or holding a passive angle. \\
    \textit{Misinterpreting Aggressive Movement} &
    High-speed movement is often labeled as \textit{flight} (rotation) when it is actually an aggressive push, or \textit{fight} (push) when it is a tactical retreat. \\
    \textit{Sniper Staticity} &
    Models consistently assume stationary snipers are "fighting," failing to recognize when they are "saving" or waiting to rotate. \\
    \textit{Utility/Knife False Negatives} &
    Models assume holding a knife or grenade always means \textit{flight}, missing instances where players use speed or utility to initiate a fight. \\
    \bottomrule
  \end{tabularx}
\end{table}

\twocolumn

\bibliographystyle{IEEEtran}
\bibliography{references_v041}

\end{document}